\documentclass[sigconf]{acmart}
\AtBeginDocument{%
  }

\setcopyright{acmlicensed}
\copyrightyear{2025}
\acmYear{2025}
\acmDOI{10.1145/3748636.3762761}  % 

\acmConference[SIGSPATIAL '25]{Proceedings of the 33rd ACM SIGSPATIAL International Conference on Advances in Geographic Information Systems}{November 03--06, 2025}{Minneapolis, MN, USA}

\acmISBN{979-8-4007-2086-4/2025/11}  % 

\usepackage[ruled,vlined]{algorithm2e}
\usepackage{adjustbox}
\usepackage{graphicx}
\usepackage{multirow}
\usepackage{adjustbox}
\usepackage{tabularx}
\usepackage{booktabs}

\newcommand{\iconleft}{\raisebox{-5pt}{\includegraphics[width=1.2em]{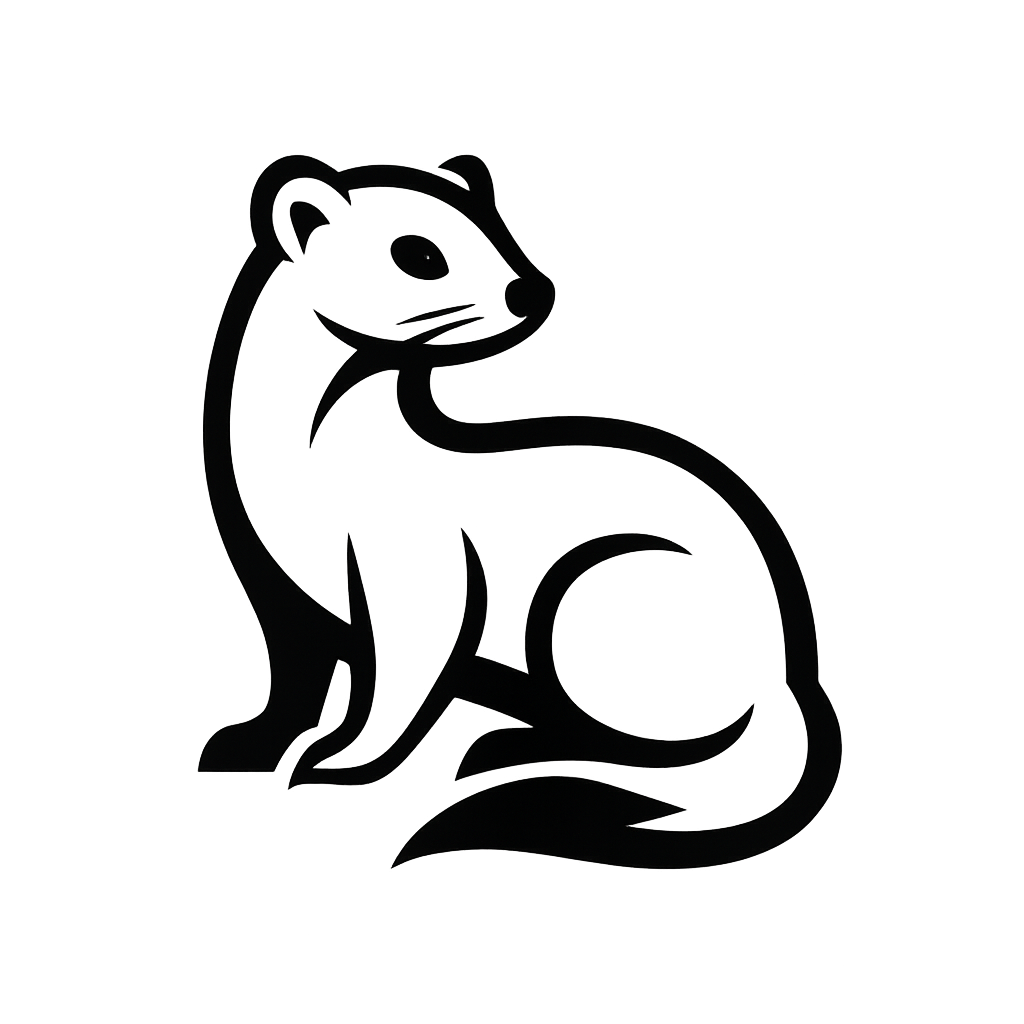}}\thinspace}

\begin{document}

%%
%% The "title" command has an optional parameter,
%% allowing the author to define a "short title" to be used in page headers.
\title{\iconleft STOAT: Spatial-Temporal Probabilistic Causal Inference Network}

%%
%% The "author" command and its associated commands are used to define
%% the authors and their affiliations.
%% Of note is the shared affiliation of the first two authors, and the
%% "authornote" and "authornotemark" commands
%% used to denote shared contribution to the research.
%\author{Ben Trovato}
%\authornote{Both authors contributed equally to this research.}
%\email{trovato@corporation.com}
%\orcid{1234-5678-9012}
%\author{G.K.M. Tobin}
%\authornotemark[1]
%\email{webmaster@marysville-ohio.com}
%\affiliation{%
%  \institution{Institute for Clarity in Documentation}
%  \city{Dublin}
%  \state{Ohio}
%  \country{USA}
%}

\author{Yang Yang}
\affiliation{%
  \institution{University of New South Wales}
  \city{Sydney}
  \state{NSW}
  \country{Australia}}
\email{yang.yang26@unsw.edu.au}

\author{Du Yin}
\affiliation{%
  \institution{University of New South Wales}
  \city{Sydney}
  \state{NSW}
  \country{Australia}}
\email{du.yin@unsw.edu.au}

\author{Hao Xue}
\affiliation{%
  \institution{University of New South Wales}
  \city{Sydney}
  \state{NSW}
  \country{Australia}}
\email{hao.xue1@unsw.edu.au}

\author{Flora Salim}
\affiliation{%
  \institution{University of New South Wales}
  \city{Sydney}
  \state{NSW}
  \country{Australia}}
\email{flora.salim@unsw.edu.au}

%%
%% By default, the full list of authors will be used in the page
%% headers. Often, this list is too long, and will overlap
%% other information printed in the page headers. This command allows
%% the author to define a more concise list
%% of authors' names for this purpose.
\renewcommand{\shortauthors}{Yang et al.}

%%
%% The abstract is a short summary of the work to be presented in the
%% article.
\begin{abstract}
Spatial-temporal causal time series (STC-TS) involve region-specific temporal observations driven by causally relevant covariates and interconnected across geographic or network-based spaces. Existing methods often model spatial and temporal dynamics independently and overlook causality-driven probabilistic forecasting, limiting their predictive power. To address this, we propose STOAT (Spatial-Temporal Probabilistic Causal Inference Network), a novel framework for probabilistic forecasting in STC-TS. The proposed method extends a causal inference approach by incorporating a spatial relation matrix that encodes interregional dependencies (e.g. proximity or connectivity), enabling spatially informed causal effect estimation. The resulting latent series are processed by deep probabilistic models to estimate the parameters of the distributions, enabling calibrated uncertainty modeling. We further explore multiple output distributions (e.g., Gaussian, Student-t, Laplace) to capture region-specific variability. Experiments on COVID-19 data across six countries demonstrate that STOAT outperforms state-of-the-art probabilistic forecasting models (DeepAR, DeepVAR, Deep State Space Model, etc.) in key metrics, particularly in regions with strong spatial dependencies. By bridging causal inference and geospatial probabilistic forecasting, STOAT offers a generalizable framework for complex spatial-temporal tasks, such as epidemic management.

\end{abstract}


%\received{20 February 2007}
%\received[revised]{12 March 2009}
%\received[accepted]{5 June 2009}

%%
%% This command processes the author and affiliation and title
%% information and builds the first part of the formatted document.
\maketitle

\section{Introduction}

\begin{figure}
    \centering
    \includegraphics[width=0.45\textwidth]{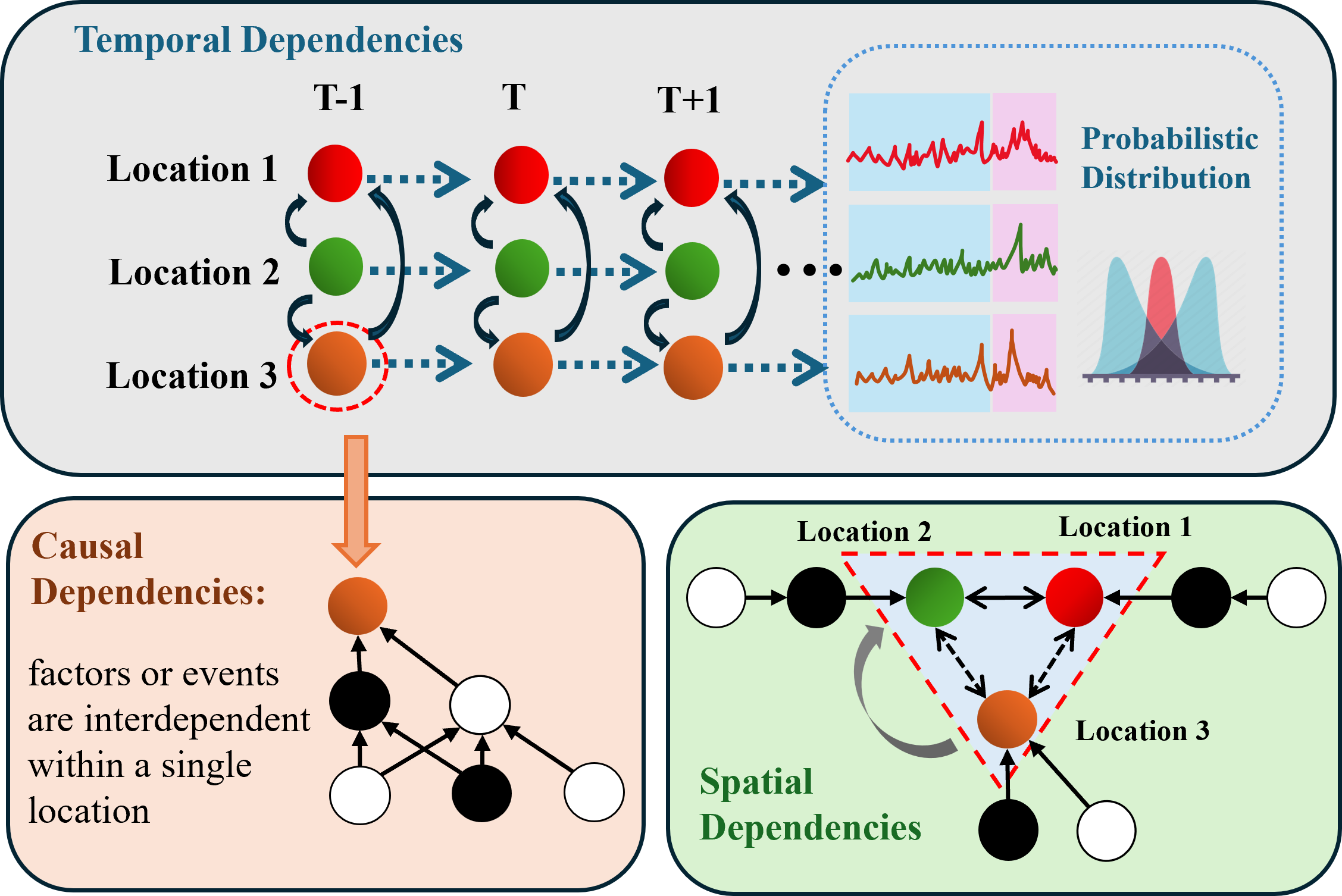}
    \caption{Illustration of the Spatial-temporal causal time series (STC-TS) setting. The bottom-left pink block (causal dependencies) depicts intra-regional causal inference, where time-varying covariates (e.g., policy interventions) influence observed outcomes within each region. Black-and-white nodes represent causally relevant covariates at specific time steps. The top gray block (temporal dependencies) shows the evolution of observed variables across three regions, with colored points (red, yellow, green) denoting distinct temporal trends. The bottom-right green block (spatial dependencies) illustrates inter-regional relationships, highlighting spatially structured influences and spillover effects. Together, the figure captures the intertwined spatial, temporal, and causal structures that define the STC-TS framework.}
    \Description{A conceptual diagram showing three interconnected blocks representing the STC-TS framework: a pink block showing causal dependencies with nodes and arrows, a gray block displaying temporal evolution across regions with colored trend lines, and a green block illustrating spatial relationships between regions.}
    \vspace{-0.2cm}
    \label{fig:IntroExample}
\end{figure}

Spatial-temporal causal time series (STC-TS) consist of regional time series that are influenced by causally relevant covariates and interconnected through spatial dependencies, as shown in Figure \ref{fig:IntroExample}. A representative case is epidemic forecasting across countries, where region-level outcomes such as case counts or hospitalizations are influenced by external factors including government interventions, vaccination rates, mobility patterns, population density, and virus evolution \cite{cao2021covid,cao2022ai,nolan2020exploring,du2025blue}. These covariates are heterogeneous, time-varying, and exhibit complex causal impacts on the temporal trajectory of the target variables. For epidemic forecasting, the target variables such as confirmed or hospitalized cases form region-specific time series, while external covariates evolve alongside, often asynchronously and irregularly. The complex causal interactions between such covariates (for instance, the joint effect of virus mutations and policy responses) remain difficult to model explicitly, particularly when considering their varying impacts across different regions \cite{mastakouri2020causal}. Furthermore, regional outcomes are not isolated phenomena. Spatial dependencies arise naturally as outbreak patterns often exhibit spillover effects due to cross-border mobility, trade connections, and latent transmission pathways \cite{wang2024learning,salim2021learning}. Understanding spatio-temporal dynamics therefore requires modeling both the causal effects of covariates within each region and the spatial influences propagating between regions. These interconnected challenges in STC-TS settings give rise to three fundamental modeling requirements: cross-region causal inference, spatial-temporal causal representation learning, and causally informed probabilistic forecasting.

\textbf{First Challenge: Cross-region causal inference.}
Factors and events across different regions (such as policy changes, mobility shifts, and vaccination progress) often have interdependent spillover effects on the outcome variables through spatial  \cite{naumann2020covid}. However, existing deep probabilistic forecasting models do not explicitly model how these cross-regional interactions and spatial dependencies jointly affect target dynamics across multiple regions. For example, DeepAR \cite{salinas2020deepar} and Deep State Space Models \cite{rangapuram2018deep} treat covariates as auxiliary inputs without capturing spatial causal relationships between regions, lacking interpretability in understanding how interventions propagate across spatial networks. Classical Difference-in-Differences (DiD) \cite{ryan2019now, metcalfe2019pay, rothbard2024tutorial} enables causal inference but is typically designed for simple treatment-control setups without considering spatial spillover effects or the complex interdependencies inherent in spatial networks. In contrast to these approaches, our framework explicitly models spatial causal mechanisms through a spatial lag term that quantifies how neighboring regions' outcomes causally influence each target region, while simultaneously incorporating treatment effects within a unified causal inference structure. Bridging this gap by developing spatially-aware causal inference frameworks that can both capture cross-regional spillover effects and provide interpretable causal estimates for complex spatio-temporal time series represents a critical yet largely unaddressed challenge in the intersection of causal inference and deep learning.

\textbf{Second Challenge: Spatial-Temporal Causal Representation Learning.}
Classical approaches in epidemiology, including compartmental models (e.g., SIR \cite{cooper2020sir}) and statistical methods such as Difference-in-Differences \cite{ohlsson2020applying, tchetgen2024universal}, Instrumental Variables (IV) \cite{baiocchi2014instrumental}, and Propensity Score Matching (PSM) \cite{kane2020propensity}, lack the capacity to capture high-dimensional, non-linear, and dynamic causal dependencies. These methods typically assume static confounding structures and fail to model complex spatio-temporal interactions in a data-driven manner. Meanwhile, existing deep temporal models often treat covariates as exogenous inputs without accounting for structural confounding or causal entanglement. Spatial dependencies, if considered at all, are rarely modeled through a causal lens. Consequently, both classical and modern approaches remain limited in their ability to infer structured causal dynamics over time and across regions. To address this, we argue that accurate forecasting in spatial-temporal causal time series (STC-TS) requires latent representations that encode both temporal dependencies and spatially heterogeneous causal effects. 

\textbf{Third Challenge: Causally Informed Probabilistic Forecasting.}
Causally informed probabilistic forecasting is critical in high-stakes scenarios like epidemic control, where uncertainty quantification is as important as point predictions \cite{xie2023overview}. However, existing methods often rely on deterministic predictions and fail to adaptively model the distributional dynamics induced by time-varying covariates or interventions, overlooking the causal relationships between covariates and targets \cite{gneiting2014probabilistic}. As a result, predictive intervals are miscalibrated and insensitive to policy changes, regional disparities, and intervention timing. Furthermore, these models typically do not distinguish between the different sources of uncertainty, such as aleatoric, epistemic, and structural uncertainty, which limits both interpretability and the robustness of downstream decision-making.

To address the unique challenges of spatial-temporal causal time series (STC-TS) forecasting, we introduce \textbf{STOAT}, a unified framework that integrates causal inference, spatial-temporal representation learning, and probabilistic forecasting. STOAT is built on the premise that reliable forecasting under covariate interventions and spatial spillovers requires more than standard temporal modeling; it requires structured causal reasoning and calibrated uncertainty estimation. The key design principle behind STOAT is to first disentangle causal effects from spatio-temporal confounding through a dedicated \textbf{Spatial-Temporal Causal Inference Module}, which extends the classical Difference-in-Differences (DiD) \cite{ryan2019now, metcalfe2019pay, rothbard2024tutorial} using a learnable spatial relation matrix. This module yields causally adjusted representations that reflect both treatment effects and spatial spillovers across regions. These representations are then fed into a \textbf{Deep Probabilistic Forecasting Module}, which captures temporal dependencies using neural sequence models and projects the latent states into structured output distributions. This enables STOAT to produce calibrated probabilistic forecasts that account for both intervention dynamics and spatial correlations. Figure~\ref{fig:ModelFramework} provides an overview of the STOAT architecture, which consists of two interconnected modules: (1) a spatial-temporal causal inference pathway that generates covariate-adjusted representations, and (2) a probabilistic forecasting pathway that learns temporal dynamics and generates distributional predictions. The primary contributions of this work are summarized as follows:

\begin{itemize}
\item \textbf{Cross-region causal inference.} 
We develop a spatially-informed causal inference mechanism that extends classical DiD using a spatial relation matrix, enabling STOAT to estimate region-specific treatment effects while accounting for spatial confounding and spillover influences.
\textit{(Addressing Challenge 1)}

\item \textbf{Causal representation learning for STC-TS.} 
By integrating causal adjustment into a neural encoder-decoder architecture, STOAT jointly models intra-regional temporal dynamics and inter-regional causal dependencies. This allows the model to learn high-dimensional spatial-temporal representations that go beyond static causal estimation.
\textit{(Addressing Challenge 2)}

\item \textbf{Causally informed probabilistic forecasting.} 
Building on causally adjusted representations, STOAT introduces structured probabilistic projection heads with support for Gaussian, Student’s-$t$, and Laplace distributions. This enables calibrated, interpretable, and region-aware uncertainty estimation that reflects intervention effects and covariate-driven dynamics.
\textit{(Addressing Challenge 3)}

\end{itemize}

\vspace{-0.1cm}
\section{Related Work}

Here, we review the related literature on causal inference for time series, deep probabilistic forecasting, and sequence modeling in the context of epidemic prediction. We highlight how existing approaches fall short in modeling spatial-temporal causal dependencies and probabilistic uncertainty jointly.

\textbf{Causal Inference in Time Series.}  
Causal inference has been extensively studied in econometrics and policy analysis, especially through the framework of Difference-in-Differences (DiD) \cite{abadie2005semiparametric, moraffah2021causal, athey2017state, runge2023causal}. DiD has been widely applied to time series data to estimate treatment effects over time by comparing treated and control groups. Recent extensions include generalized DiD methods for staggered treatment adoption \cite{sun2021estimating} and synthetic control-based approaches \cite{abadie2010synthetic}. However, these methods are typically restricted to one-dimensional or pairwise group comparisons, and do not account for spatial interactions across regions. They also assume static treatment effects and rarely integrate with representation learning. While some recent work incorporates DiD with neural networks \cite{zhou2024machine, schwab2020learning}, they lack structured spatial modeling and do not support multi-region causal adjustment. Our work addresses this gap by integrating spatial causal inference into a deep temporal forecasting framework using a structured adjacency matrix for regional spillover modeling.

\textbf{Deep Probabilistic Time-Series Forecasting.}  
Several deep learning models have been proposed for probabilistic time series forecasting by combining recurrent neural networks with statistical likelihood estimation. DeepAR \cite{salinas2020deepar} and MTSNet \cite{yang2023mtsnet} model each time series independently under a univariate distribution, using shared RNN parameters to learn sequential dependencies. Deep State Space Models (DSSM) \cite{rangapuram2018deep} incorporate a linear state-space formulation with recurrent updates, but similarly assume independent marginal distributions for each series. DeepVAR \cite{salinas2019high} enhances this with Gaussian copula processes to capture dependencies between target variables but does not model structured causal relationships across space or time. DeepFactor \cite{wang2019deep} decomposes global and local components using latent variables, yet still treats covariates as exogenous inputs without causal interpretation. None of these models incorporate causal inference mechanisms or address confounding in the presence of spatially distributed covariates. In contrast, STOAT explicitly adjusts for covariate-driven biases through spatial causal representations and generates calibrated predictive distributions informed by intervention dynamics.

\textbf{Sequence Modeling for Epidemic Forecasting.}  
Sequence-to-sequence (seq2seq) models and encoder-decoder architectures have been widely applied in epidemic forecasting \cite{pathan2020time, alassafi2022time, shahid2020predictions}. These models, including GRU-based and transformer variants, are primarily trained for point prediction, optimizing loss functions such as MSE or MAE. While effective in short-term forecasting, they do not quantify predictive uncertainty or incorporate causal reasoning. Classical epidemiological models, such as SIR and SEIR variants \cite{cao2021covid, he2020seir, carcione2020simulation,lopez2021modified}, provide mechanistic insights into disease transmission, but they rely on fixed parametric assumptions and are difficult to scale or integrate with heterogeneous covariates. Moreover, these models often do not support flexible uncertainty modeling or data-driven representation learning. Our work bridges this gap by combining causal inference with probabilistic deep learning. STOAT allows for region-aware, covariate-adjusted forecasting with interpretable uncertainty, offering a unified approach to modeling spatial-temporal dynamics in complex real-world settings.
Compared to existing approaches, STOAT differs in three key aspects:
\begin{itemize}
\item[(1)] \textbf{Spatial Causal Adjustment:} STOAT explicitly incorporates causal inference into deep forecasting by extending Difference-in-Differences (DiD) with a spatial relation matrix, enabling causal adjustment across interconnected regions—something not supported by classical DiD or deep forecasting models.
\item[(2)] \textbf{Causal Representation Learning:} Unlike existing deep probabilistic models (e.g., DeepAR, DeepVAR), which treat covariates as exogenous and non-causal, STOAT constructs causally adjusted latent representations that reflect both spatial spillovers and intervention effects over time.
\item[(3)] \textbf{Causally Informed Probabilistic Forecasting:} While prior deep forecasting models often estimate point values or assume fixed likelihoods, STOAT performs distributional prediction by estimating the parameters of output distributions (e.g., Gaussian, Student's-$t$, Laplace), conditioned on causally disentangled temporal dynamics.
\end{itemize}

\section{STOAT: Spatial-Temporal Probabilistic Causal Inference Network}

\subsection{Overview of STOAT}

%STOAT tackles STC-TS forecasting with a Spatial-Temporal Causal Inference Module for region-specific causal adjustment, followed by a Deep Probabilistic Projection Module for temporal dynamics and distributional forecasting (Gaussian, Student's-$t$, Laplace), as shown in Figure \ref{fig:ModelFramework}.

\begin{figure*}[!t]
    \centering
    \includegraphics[width=1\textwidth]{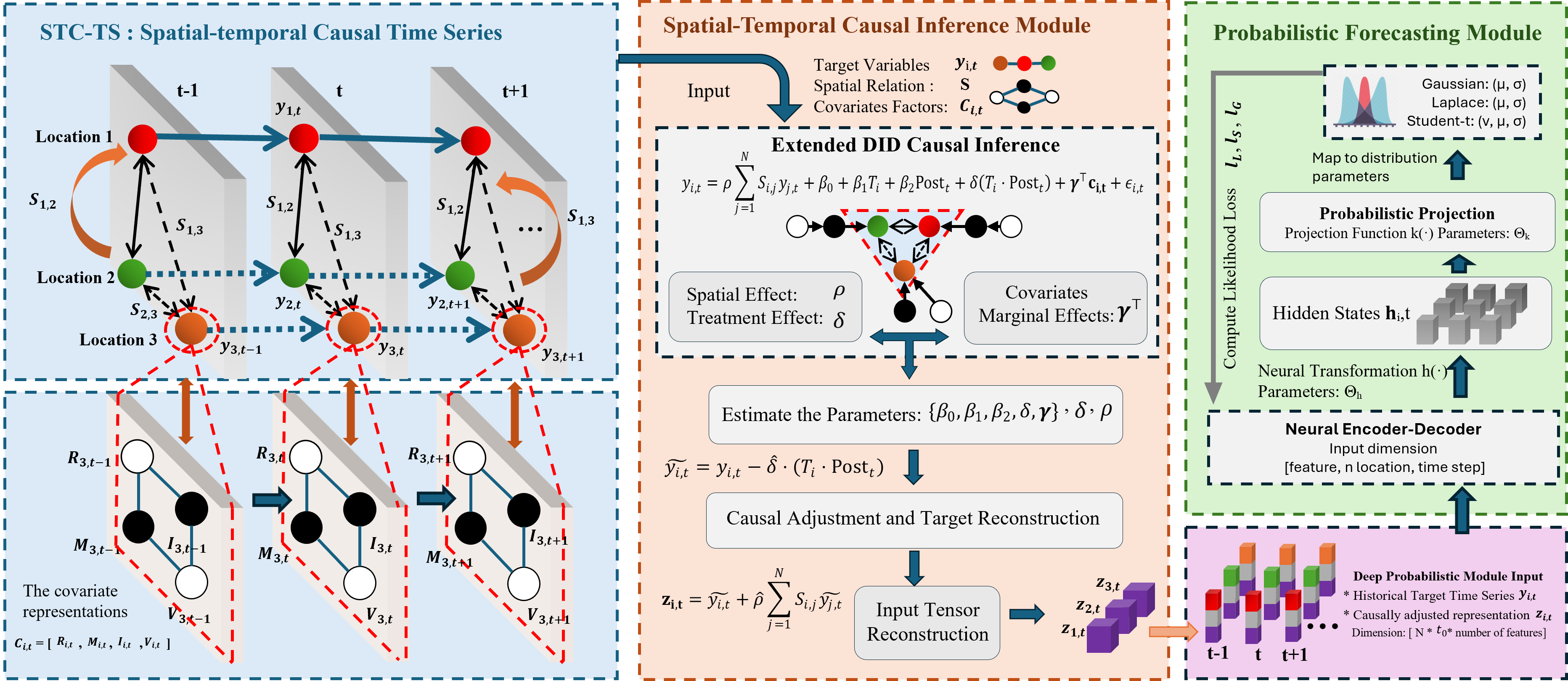}
    \caption[STOAT Architecture Overview]{Overview of the STOAT architecture, which consists of two interconnected modules: a spatial-temporal causal inference pathway that generates covariate-adjusted representations, and a probabilistic forecasting pathway that learns temporal dynamics and generates distributional predictions. In what follows, we detail each component of the STOAT framework.}
    \label{fig:ModelFramework}
\end{figure*}

To solve the STC-TS probabilistic forecasting problem, STOAT employs two key modules as shown in Figure \ref{fig:ModelFramework}. The \textbf{Spatial-Temporal Causal Inference Module} first performs causally-informed representation learning by extending the Difference-in-Differences framework with spatial relation matrices, enabling region-aware causal adjustment that disentangles treatment effects from confounding factors across interconnected spatial units. The causally adjusted latent representations are then fed into the \textbf{Deep Probabilistic Projection Module}, which employs neural encoder-decoder architectures to capture complex temporal dependencies and generates distributional forecasts by estimating parameters of flexible output distributions (Gaussian, Student's-$t$, Laplace), thereby providing calibrated uncertainty quantification that reflects both spatial spillover effects and intervention-driven dynamics in the forecasting process.

Spatial-temporal causal time series (STC-TS) consisting of target variables time-series \(\mathbf{y}\) and multiple causally relevant covariates time-series \(\mathbf{C}\) exhibit both inter-regional spatial dependencies and intra-regional temporal couplings, as shown in \figurename~\ref{fig:IntroExample}. For the considered epidemic forecasting data, we assume \(\mathbf{y}_{t}\) represents a target vector composed of the values of the multivariate time series at time step \(t\), e.g., confirmed cases from \(N\) regions/countries. \(\mathbf{C}_{t}\) represents the causally relevant covariates corresponding to the target multivariate time series, e.g., policy stringency index \cite{hale2021global,gros2021new}, vaccination rates, ICU patient and reproduction number \cite{alimohamadi2020estimate} from \(N\) regions/countries, where these covariates serve as causal factors (treatments) that influence the target outcomes. STOAT characterizes the STC-TS probabilistic forecasting problem in Eq. (\ref{eq:stc_ts}),
\begin{equation}
P\left(\mathbf{y}_{t_{0}+1:t_{0}+m} \mid \mathbf{y}_{1:t_{0}}, \mathbf{C}_{1:t_{0}+m}, \mathbf{S}, \mathbf{\Theta}\right),
\label{eq:stc_ts}
\end{equation}
where \(\mathbf{\Theta}\) denotes the parameters of STOAT, \(\mathbf{S}\) is the spatial relation matrix capturing inter-regional spillover effects and confounding relationships, the historical time series steps are \(\{1, 2, \ldots, t_{0}\}\), and the prediction time length is \(m \in \mathbb{N}\). STOAT models the conditional joint multivariate probability distributions of the future multivariate time series \(\mathbf{y}_{t_{0}+1:t_{0}+m}\), given the historical multivariate time series \(\mathbf{y}_{1:t_{0}}\), their causally relevant covariates \(\mathbf{C}_{1:t_{0}+m}\), and the spatial relation matrix \(\mathbf{S}\), which is incorporated within the model architecture to account for inter-regional dependencies during the causal adjustment and forecasting processes.

\subsection{Spatial-Temporal Causal Inference Module}
The Spatial-Temporal Causal Inference Module performs causally-informed representation learning by extending the Difference-in-Differences \cite{moraffah2021causal} framework with spatial relation matrices, enabling region-aware causal adjustment representation learning.

\subsubsection{Spatial Relation Matrix Construction}
Given the geographic coordinates $(\text{lat}_i, \text{lon}_i)$ for each region $i$, we construct a spatial relation matrix $\mathbf{S} \in \mathbb{R}^{N \times N}$ to capture the spatial dependencies between regions. We first compute the geodesic distance $d_{i,j}$ between each pair of regions $i$ and $j$, then define the spatial relation matrix as:
\begin{equation}
\mathbf{S} = 
\begin{bmatrix}
0 & S_{1,2} & \cdots & S_{1,N} \\
S_{2,1} & 0 & \cdots & S_{2,N} \\
\vdots & \vdots & \ddots & \vdots \\
S_{N,1} & S_{N,2} & \cdots & 0
\end{bmatrix},
\quad
S_{i,j} = \begin{cases}
\frac{ \frac{1}{d_{i,j}^{\alpha}} }{ \sum_{k \neq i} \frac{1}{d_{i,k}^{\alpha}} } & \text{if } i \neq j \\
0 & \text{if } i = j
\end{cases}
\end{equation}
The parameter $\alpha > 0$ controls the spatial decay rate, with higher values of $\alpha$ leading to more localized spatial effects. The row-normalization ensures that $\sum_{j=1}^{N} S_{i,j} = 1$ for all $i$, making $\mathbf{S}$ a row-stochastic matrix that preserves the interpretation of spatial weights as proportional influences.

While we focus on distance-based relationships in this formulation, the spatial relation matrix $\mathbf{S}$ can be constructed using alternative proximity measures such as economic connectivity, transportation networks, administrative boundaries, or hybrid approaches that combine multiple spatial relationship indicators.

\subsubsection{Spatial Causal Inference}

The spatial causal inference mechanism in STOAT extends the classical Difference-in-Differences (DiD) framework by incorporating spatial dependencies through the spatial relation matrix $\mathbf{S}$ constructed in Section 3.2.1. This spatially-informed approach enables region-aware causal adjustment that accounts for both direct treatment effects and spatial spillover influences across interconnected regions.

\textbf{Spatially-Enhanced DiD Formulation.}
To capture the complex interplay between spatial dependencies and causal effects in epidemic dynamics, we formulate the core spatial causal inference model as:
\begin{equation}
y_{i,t} = \rho \sum_{j=1}^{N} S_{i,j} y_{j,t-1} + \beta_0 + \beta_1 T_i + \beta_2 \text{Post}_t + \delta (T_i \cdot \text{Post}_t) + \boldsymbol{\gamma}^{\top} \mathbf{c}_{i,t} + \epsilon_{i,t}
\label{eq:spatial_did}
\end{equation}

where $y_{i,t}$ represents the observed outcome (e.g., COVID-19 cases) for region $i$ at time $t$. The spatial lag term $\sum_{j=1}^{N} S_{i,j} y_{j,t-1}$ captures the weighted average of neighboring regions' outcomes based on the spatial relation matrix, with coefficient $\rho$ controlling the strength of spatial spillover effects. The treatment indicator $T_i$ is a binary variable denoting whether region $i$ has adopted the intervention, while $\text{Post}_t$ indicates the post-treatment period. The parameter $\delta$ represents the treatment effect, capturing the causal impact of interventions on target outcomes.

\subsubsection{Multi-dimensional Covariates Construction}
Building upon the spatially-enhanced DiD framework in Eq. (\ref{eq:spatial_did}), we incorporate epidemiological covariates through a structured multi-dimensional vector. The covariate vector $\mathbf{c}_{i,t}$ encompasses four key epidemiological dimensions:
\begin{equation}
\mathbf{c}_{i,t} = \begin{bmatrix} R_{i,t} \\ M_{i,t} \\ V_{i,t} \\ I_{i,t} \end{bmatrix}, \quad \boldsymbol{\gamma} = \begin{bmatrix} \gamma_1 \\ \gamma_2 \\ \gamma_3 \\ \gamma_4 \end{bmatrix}
\label{eq:covariate_structure}
\end{equation}
where $R_{i,t}$ denotes the Reproduction Number capturing transmission dynamics, $M_{i,t}$ represents the Mitigation Stringency Index measuring policy intervention intensity, $V_{i,t}$ indicates the Vaccination Share representing immunization coverage, and $I_{i,t}$ represents the ICU Patient Number indicating healthcare system strain. The corresponding parameter vector $\boldsymbol{\gamma}$ captures the marginal effects of each covariate dimension.

The covariate contribution in Eq. (\ref{eq:spatial_did}) is explicitly modeled as:
\begin{equation}
\boldsymbol{\gamma}^{\top} \mathbf{c}_{i,t} = \gamma_1 R_{i,t} + \gamma_2 M_{i,t} + \gamma_3 V_{i,t} + \gamma_4 I_{i,t}
\label{eq:covariate_effects}
\end{equation}
This formulation allows for the simultaneous estimation of individual covariate effects while maintaining the spatial-temporal structure inherent in the epidemic forecasting problem.

\subsubsection{Parameter Estimation and Identification}
The estimation of parameters in Eq. (\ref{eq:spatial_did}) follows a two-stage approach. First, we estimate the spatial autoregressive coefficient $\rho$ using instrumental variable techniques to address endogeneity in the spatial lag term. Subsequently, we apply ordinary least squares (OLS) to estimate the remaining parameters $\{\beta_0, \beta_1, \beta_2, \delta, \boldsymbol{\gamma}\}$, given the estimated $\hat{\rho}$. The identification of $\delta$ relies on the parallel trends assumption, which requires that treated and control regions would have followed similar trajectories in the absence of treatment.

\subsubsection{Causal Adjustment and Target Reconstruction}
Following the parameter estimation from Eq. (\ref{eq:spatial_did}), STOAT performs causal adjustment by constructing counterfactual outcomes that isolate the underlying temporal dynamics from intervention effects. The causally adjusted target variable is obtained as:
\begin{equation}
\tilde{y}_{i,t} = y_{i,t} - \hat{\delta} \cdot (T_i \cdot \text{Post}_t)
\label{eq:causal_adjustment}
\end{equation}
where $\hat{\delta}$ represents the estimated treatment effect from Eq. (\ref{eq:spatial_did}). This adjustment yields causally corrected target variables $\tilde{y}_{i,t}$ that represent the counterfactual outcomes in the absence of treatment effects, thereby isolating the underlying epidemiological dynamics from policy-driven changes.

To construct the input tensor for the subsequent deep probabilistic forecasting module, we combine the causally adjusted targets with the spatial lag effects:
\begin{equation}
\mathbf{z}_{i,t} = \tilde{y}_{i,t} + \hat{\rho} \sum_{j=1}^{N} S_{i,j} \tilde{y}_{j,t}
\label{eq:spatial_adjusted_input}
\end{equation}

The reconstructed variable $\tilde{y}_{i,t}$ and spatially adjusted input $\mathbf{z}_{i,t}$ explicitly incorporate information from the spatial relation matrix $\mathbf{S}$ through spatial lag effects and the causally relevant covariates $\mathbf{c}$ through causal adjustments, effectively capturing the influence of spatial dependencies and covariate-driven interventions.

Thus, for subsequent deep probabilistic forecasting, we leverage $\mathbf{z}_{i,t}$ directly, encapsulating the information of spatial dependencies and covariate-driven causal effects. Therefore, the conditional probability formulation can be simplified from its original form:
\begin{equation}
P\left(\mathbf{y}_{t_{0}+1:t_{0}+m} \mid \mathbf{y}_{1:t_{0}}, \mathbf{C}_{1:t_{0}+m}, \mathbf{S}, \boldsymbol{\Theta}\right)
\label{eq:prob_forecast_original}
\end{equation}
to the simplified form:
\begin{equation}
P\left(\mathbf{y}_{t_{0}+1:t_{0}+m} \mid \mathbf{y}_{1:t_{0}} , \mathbf{z}_{i, 1:t_{0}}, \boldsymbol{\Theta}\right)
\label{eq:prob_forecast_simplified1}
\end{equation}
where $\boldsymbol{\Theta}$ represents the parameters of the deep probabilistic neural network shared across all regions.

\subsubsection{Interpretability and Causal Parameters}
The spatial causal inference mechanism provides three key interpretable parameters that offer insights into the underlying epidemic dynamics:

The treatment effect parameter $\delta$ quantifies the average causal impact of interventions on target outcomes across regions. A significant negative $\delta$ (e.g., $\delta < -0.1$) indicates that policy interventions effectively reduce case counts, while $\delta \approx 0$ suggests limited intervention effectiveness.

The covariate effect vector $\boldsymbol{\gamma}$ from Eq. (\ref{eq:covariate_effects}) captures the marginal contribution of each epidemiological factor. For instance, $\gamma_1 > 0$ confirms that higher reproduction numbers drive increased transmission, while $\gamma_2 < 0$ validates that stricter policies reduce case growth rates. The magnitude of $|\gamma_3|$ reflects the relative importance of vaccination coverage in epidemic control.

The spatial spillover parameter $\rho$ measures inter-regional dependency strength. Values of $\rho > 0.3$ indicate strong positive spatial correlation, suggesting that neighboring regions' epidemic patterns significantly influence local outcomes through cross-border mobility or policy coordination effects.

\subsubsection{Integration with Deep Probabilistic Forecasting}
The causally adjusted spatial representations $\mathbf{z}_{i,t}$ from Eq. (\ref{eq:spatial_adjusted_input}) serve as inputs to the deep probabilistic projection module described in Section 3.3. This integration ensures that subsequent probabilistic forecasting operates on causally disentangled temporal patterns while preserving essential spatial dependencies. The framework enables STOAT to generate predictions that reflect both underlying epidemiological dynamics and expected intervention effects, providing a principled foundation for uncertainty quantification in policy-sensitive forecasting scenarios.
\subsection{Deep Probabilistic Forecasting Module}

The deep probabilistic forecasting module in STOAT leverages the causally adjusted spatial-temporal representations $\mathbf{z}_{i,t}$ derived from the spatial causal inference module (Eq. (\ref{eq:spatial_adjusted_input})). These representations encapsulate both intra-regional temporal dynamics and inter-regional spatial dependencies, effectively integrating the effects of the causally relevant covariates $\mathbf{c}_{i,t}$ and the spatial relation matrix $\mathbf{S}$. By using $\mathbf{z}_{i,t}$, the forecasting problem is simplified to model the conditional joint probability distribution of future multivariate time series $\mathbf{y}_{t_{0}+1:t_{0}+m}$, given the historical target series $\mathbf{y}_{1:t_{0}}$ and the causally adjusted representations $\mathbf{z}_{i,1:t_{0}}$. The probabilistic forecasting problem is formulated as:
\begin{equation}
P\left(\mathbf{y}_{t_{0}+1:t_{0}+m} \mid \mathbf{y}_{1:t_{0}}, \mathbf{z}_{i,1:t_{0}}, \mathbf{\Theta}\right),
\label{eq:prob_forecast_simplified}
\end{equation}
where $\mathbf{\Theta}$ denotes the learnable parameters of STOAT's neural architecture, the historical time steps span $\{1, 2, \ldots, t_{0}\}$, and the prediction horizon is $m \in \mathbb{N}$. The use of $\mathbf{z}_{i,1:t_{0}}$ as a conditioning variable simplifies the model by embedding the spatial and causal effects into a unified representation, enabling robust forecasting that accounts for causal interventions and spillover effects.

Assuming the time series points are conditionally independent given the latent representations, the joint distribution in Eq. (\ref{eq:prob_forecast_simplified}) can be factorized as:
\begin{equation}
\begin{aligned}
P\left(\mathbf{y}_{t_{0}+1:t_{0}+m} \mid \mathbf{y}_{1:t_{0}}, \mathbf{z}_{i,1:t_{0}}, \mathbf{\Theta}\right) = \\
\prod_{t=t_{0}}^{t_{0}+m-1} P\left(\mathbf{y}_{t+1} \mid \mathbf{y}_{1:t}, \mathbf{z}_{i,1:t}, \mathbf{\Theta}\right).
\end{aligned}
\label{eq:factorized_prob}
\end{equation}

To model this conditional distribution, STOAT employs a neural encoder-decoder architecture that processes the causally adjusted representations $\mathbf{z}_{i,t}$. Specifically, the latent representation at time $t$ for region $i$, denoted $\mathbf{h}_{i,t}$, is generated by a neural transformation function $h(\cdot)$ that captures temporal dependencies:
\begin{equation}
\mathbf{h}_{i,t} = h\left(\mathbf{z}_{i,1:t}, \mathbf{y}_{1:t}, \mathbf{h}_{i,t-1}, \mathbf{\Theta}_{h}\right),
\label{eq:latent_h}
\end{equation}
where $\mathbf{z}_{i,1:t}$ is the sequence of causally adjusted representations for region $i$ up to time $t$, $\mathbf{y}_{1:t}$ includes the historical target series, $\mathbf{h}_{i,t-1}$ represents the hidden state from the previous time step, and $\mathbf{\Theta}_{h}$ denotes the parameters of the neural transformation. The function $h(\cdot)$ is implemented using a stacked recurrent neural network (e.g., LSTM or GRU) to capture long-term temporal dependencies, leveraging the embedded spatial and causal information in $\mathbf{z}_{i,1:t}$.

The conditional probability in Eq. (\ref{eq:factorized_prob}) is modeled as a product of likelihood functions, where the parameters of the output distribution are determined by a projection layer $k(\cdot)$. For each region $i$ and time $t+1$, the likelihood is defined as:
\begin{equation}
\begin{aligned}
\prod_{t=t_{0}}^{t_{0}+m-1} P\left(\mathbf{y}_{t+1} \mid \mathbf{y}_{1:t}, \mathbf{z}_{i,1:t}, \mathbf{\Theta}\right) = \\
\prod_{t=t_{0}}^{t_{0}+m-1} \ell\left(y_{i,t+1} \mid k\left(\mathbf{h}_{i,t}, \mathbf{\Theta}_{k}\right)\right),
\end{aligned}
\label{eq:likelihood_product}
\end{equation}
where $\ell(\cdot)$ is the likelihood function, and $k(\cdot)$ maps the latent representation $\mathbf{h}_{i,t}$ to the parameters of the output distribution, with learnable parameters $\mathbf{\Theta}_{k}$. The global parameter set $\mathbf{\Theta}$ comprises $\mathbf{\Theta}_{h}$ and $\mathbf{\Theta}_{k}$.

\subsubsection{Probabilistic Projection}

To capture the diverse statistical properties of STC-TS data (e.g., epidemic case counts), STOAT supports multiple output distributions, including Gaussian, Laplace, and Student's-$t$, to model region-specific variability and uncertainty. The projection layer $k(\cdot)$ maps the latent representation $\mathbf{h}_{i,t}$ to the parameters of these distributions. The likelihood functions and corresponding projection mappings are defined as follows:

Laplace distribution likelihood function:
\begin{equation}
\ell_{L}(y_{i,t} \mid \mu, \sigma) = \frac{1}{2\sigma} \exp\left(-\frac{|y_{i,t} - \mu|}{\sigma}\right),
\label{eq:laplace}
\end{equation}

Gaussian distribution likelihood function:
\begin{equation}
\ell_{G}(y_{i,t} \mid \mu, \sigma) = \frac{1}{\sqrt{2\pi\sigma^{2}}} \exp\left(-\frac{(y_{i,t} - \mu)^{2}}{2\sigma^{2}}\right),
\label{eq:gaussian}
\end{equation}

Student's-$t$ distribution likelihood function:
\begin{equation}
\ell_{S}(y_{i,t} \mid \nu, \mu, \sigma) = \frac{\Gamma\left(\frac{\nu+1}{2}\right)}{\Gamma\left(\frac{\nu}{2}\right)\sqrt{\pi\nu}\sigma} \left(1 + \frac{1}{\nu}\left(\frac{y_{i,t} - \mu}{\sigma}\right)^{2}\right)^{-\frac{\nu+1}{2}},
\label{eq:student_t}
\end{equation}

\subsubsection{Likelihood Loss}

To optimize the parameters $\mathbf{\Theta} = \{\mathbf{\Theta}_{h}, \mathbf{\Theta}_{k}\}$, STOAT employs a negative log-likelihood loss function, defined as:
\begin{equation}
\mathcal{L} = -\sum_{i=1}^{N} \sum_{t=t_{0}}^{t_{0}+m-1} \log \ell\left(y_{i,t+1} \mid k\left(\mathbf{h}_{i,t}, \mathbf{\Theta}_{k}\right)\right).
\label{eq:loss}
\end{equation}
This loss is minimized using stochastic gradient descent, enabling the model to learn parameters that maximize the likelihood of the observed data under the specified distribution.

During training, both the conditioning range $\{1, 2, \ldots, t_{0}\}$ and the prediction range $\{t_{0}+1, \ldots, t_{0}+m\}$ are available. For prediction, only the historical data $y_{i,t}$ ($t \leq t_{0}$) and causally adjusted representations $\mathbf{z}_{i,1:t_{0}}$ are used as inputs. The forecasted values $\tilde{y}_{i,t}$ ($t > t_{0}$) are sampled from the estimated distribution:
\begin{equation}
\tilde{\mathbf{y}}_{i,t_{0}+1:t_{0}+m} \sim P\left(\mathbf{y}_{i,t_{0}+1:t_{0}+m} \mid \mathbf{y}_{i,1:t_{0}}, \mathbf{z}_{i,1:t_{0}}, \mathbf{\Theta}\right).
\label{eq:sampling}
\end{equation}
These samples are generated via ancestral sampling, where each $\tilde{y}_{i,t+1}$ is drawn from the distribution parameterized by $k(\mathbf{h}_{i,t}, \mathbf{\Theta}_{k})$ and used as input for the next time step until $t = t_{0}+m$. This process enables STOAT to produce probabilistic forecasts with calibrated uncertainty, leveraging the causally adjusted representations $\mathbf{z}_{i,t}$ to reflect both spatial dependencies and intervention effects.

The overall algorithm is shown in Appendix~\ref{Algorithm}

\section{Experimental Evaluation}
To assess the performance of the proposed STOAT framework on spatial-temporal causal multivariate time series (STC-MTS) data, we conduct experiments using a comprehensive COVID-19 dataset. We compare STOAT against state-of-the-art probabilistic time-series forecasting methods, focusing on its ability to model inter-regional spatial dependencies and causal effects.
\subsection{Experimental Setup}
\subsubsection{COVID-19 Dataset with Spatial and Causal Factors}
The experiments utilize a COVID-19 dataset comprising target time series and causally relevant factors from six countries across different continents: UK, US, Canada, France, Spain, and Italy. These countries were selected due to their diverse epidemic trajectories, varied non-pharmaceutical intervention (NPI) policies, and geographic distribution across multiple continents, enabling evaluation of STOAT's generalization across different spatial and temporal contexts.
\begin{itemize}
\item \textit{Target time series:} The dataset includes daily confirmed COVID-19 cases, sourced from the Johns Hopkins University COVID-19 data repository. Data spans from June 1, 2020, to June 30, 2022, covering a period of significant epidemic variability.
\item \textit{Causal Factors:} The causal factors include ICU admissions, vaccination booster rates, Stringency Index \cite{hale2021global}, and daily reproduction number, representing virus mutation. These factors constitute the covariate matrix $\mathbf{C}$. Omicron proportion data is sourced from the GISAID initiative\footnote{Available at \href{https://covariants.org/}{covariants.org}.}, vaccination data from Our World in Data\footnote{Available at \href{https://github.com/owid/covid-19-data/tree/master/public/data/vaccinations}{github.com/owid/covid-19-data}.}, Stringency Index from the Oxford Covid-19 Government Response Tracker, and reproduction rates from a public repository \cite{abadie2005semiparametric}\footnote{Available at \href{https://github.com/crondonm/TrackingR}{github.com/crondonm/TrackingR}.}.
\item \textit{Spatial dependencies:} Inter-regional couplings are modeled using the spatial relation matrix $\mathbf{S}$, constructed from geodesic distances between countries' geographic coordinates (latitude and longitude).
\end{itemize}
The dataset from June 1, 2020 is used for training, with the goal of predicting daily confirmed cases and ICU admissions over forecast horizons of 5, 7, and 10 days. Due to the unique combination of spatial dependencies and multiple causal factors in this dataset, no other comparable time-series datasets are currently available for evaluation.

\subsubsection{Baseline Models}
We compare STOAT with leading probabilistic time-series forecasting models implemented in the GluonTS toolkit \cite{alexandrov2020gluonts}: DeepAR, Deep State Space Model (DSSM), DeepFactor (DF), MTSNet \cite{yang2023mtsnet}, and DeepVAR. Non-probabilistic methods (e.g., ARIMA, LSTM, etc.) are excluded, as they do not produce probabilistic forecasts suitable for comparison. As STOAT focuses on probabilistic forecasting to evaluate distribution accuracy, we do not compare with point-based spatio-temporal forecasting models.
\begin{itemize}
\item \textit{DeepAR} \cite{salinas2020deepar}: Utilizes a two-layer RNN to extract temporal features, mapping hidden states to distribution parameters for probabilistic predictions.
\item \textit{DSSM} \cite{rangapuram2018deep}: Combines a linear state-space model with RNNs, incorporating Gaussian noise for tractable likelihood-based training.
\item \textit{DF} \cite{wang2019deep}: Employs RNNs for global time-series representations, using Kalman filtering to generate Gaussian-based probabilistic forecasts.
\item \textit{MTSNet} \cite{yang2023mtsnet}: Leverages global learning framework with RNNs to model multiple time series, outputting parameters for probabilistic distributions.
\item \textit{DeepVAR} \cite{salinas2019high}: Extends vector autoregression with deep learning, using RNNs to model multivariate dependencies with Copula function and produce probabilistic forecasts.
\end{itemize}
To evaluate the contributions of STOAT's spatial causal inference and causal factor integration, we test the following variants:
\begin{itemize}
\item \textit{STOAT-Laplace}: Uses Laplace distribution with full spatial causal adjustment (via $\mathbf{S}$) and causal factors $\mathbf{C}$.
\item \textit{STOAT-Laplace-NoSpatial}: Uses Laplace distribution without spatial causal adjustment (i.e., bypassing $\mathbf{S}$).
\item \textit{STOAT-Laplace-NoFactors}: Uses Laplace distribution without causal factors $\mathbf{C}$.
\item \textit{STOAT-Gaussian}: Uses Gaussian distribution with full spatial causal adjustment and causal factors.
\item \textit{STOAT-Gaussian-NoSpatial}: Uses Gaussian distribution without spatial causal adjustment.
\item \textit{STOAT-Gaussian-NoFactors}: Uses Gaussian distribution without causal factors.
\item \textit{STOAT-StudentT}: Uses Student's-$t$ distribution with full spatial causal adjustment and causal factors.
\item \textit{STOAT-StudentT-NoSpatial}: Uses Student's-$t$ distribution without spatial causal adjustment.
\item \textit{STOAT-StudentT-NoFactors}: Uses Student's-$t$ distribution without causal factors.
\end{itemize}

\begin{table*}[!ht]
    \centering
    \renewcommand{\arraystretch}{1.0} % Adjust the row height factor
    \footnotesize % Adjust font size
    \adjustbox{max width=\textwidth}{
    \begin{tabularx}{\textwidth}{l|l|*{8}{>{\centering\arraybackslash}X}}
    \toprule
    Time Period & Metrics & DeepVAR & DeepAR & DSSM & DeepFactor & MTSNet & STOAT-Student-T & STOAT-Gaussian & STOAT-Laplace \\
    \midrule
    \multirow{8}{*}{\textbf{5 Days}} & Quantile Loss [0.1] & 0.1894 & 0.1723 & 0.2015 & 0.1976 & 0.1845 & 0.1543 & \textbf{0.1421} & 0.1478 \\
    & Quantile Loss [0.5] & 0.3742 & \textbf{0.3421} & 0.3927 & 0.3814 & 0.3598 & 0.3476 & 0.3378 & 0.3312 \\
    & Quantile Loss [0.9] & 0.2653 & 0.2512 & 0.2789 & 0.2478 & 0.2634 & \textbf{0.2356} & 0.2412 & 0.2389 \\
    & Coverage [0.1] & 0.2000 & 0.3000 & 0.4000 & 0.1000 & 0.2000 & 0.1000 & \textbf{0.1000} & 0.1000 \\
    & Coverage [0.5] & 0.4000 & 0.6000 & 0.3000 & 0.7000 & \textbf{0.5000} & 0.4000 & 0.5000 & 0.5000 \\
    & Coverage [0.9] & 0.8000 & \textbf{0.9000} & 0.6000 & 0.8000 & 0.7000 & \textbf{0.9000} & 0.8000 & \textbf{0.9000} \\
    & CRPS & 0.3298 & 0.2954 & 0.3512 & 0.3176 & 0.3289 & 0.2897 & \textbf{0.2745} & 0.2798 \\
    & Energy Score & 1.2789 & 1.1432 & 1.4123 & 1.3214 & 1.2987 & 1.1012 & \textbf{1.0789} & 1.0876 \\
    \midrule
    \multirow{8}{*}{\textbf{7 Days}} & Quantile Loss [0.1] & 0.2187 & 0.2056 & 0.2312 & 0.2234 & 0.2112 & 0.1898 & \textbf{0.1789} & 0.1845 \\
    & Quantile Loss [0.5] & 0.5245 & 0.4876 & 0.5467 & 0.5321 & 0.4987 & 0.4723 & 0.4612 & \textbf{0.4534} \\
    & Quantile Loss [0.9] & 0.2812 & 0.2678 & 0.2934 & 0.2712 & \textbf{0.2545} & 0.2612 & 0.2589 & 0.2567 \\
    & Coverage [0.1] & 0.3000 & 0.2000 & 0.4000 & 0.1000 & 0.2000 & 0.1000 & \textbf{0.1000} & 0.1000 \\
    & Coverage [0.5] & 0.4000 & 0.6000 & 0.3000 & 0.7000 & \textbf{0.5000} & 0.4000 & 0.5000 & 0.5000 \\
    & Coverage [0.9] & 0.7000 & \textbf{0.9000} & 0.6000 & 0.8000 & 0.8000 & \textbf{0.9000} & 0.8000 & \textbf{0.9000} \\
    & CRPS & 0.3623 & 0.3478 & 0.3845 & 0.3712 & 0.3543 & 0.3176 & \textbf{0.3056} & 0.3123 \\
    & Energy Score & 1.4678 & 1.3765 & 1.5345 & 1.4567 & 1.4321 & 1.2234 & \textbf{1.1876} & 1.1987 \\
    \midrule
    \multirow{8}{*}{\textbf{10 Days}} & Quantile Loss [0.1] & 0.2987 & 0.2765 & 0.3123 & 0.2876 & 0.2934 & 0.2567 & \textbf{0.2456} & 0.2512 \\
    & Quantile Loss [0.5] & 0.7234 & 0.6987 & 0.7456 & 0.7321 & 0.7123 & 0.6654 & 0.6456 & \textbf{0.6234} \\
    & Quantile Loss [0.9] & 0.3789 & 0.3567 & 0.3923 & 0.3678 & 0.3545 & 0.3321 & 0.3289 & \textbf{0.3212} \\
    & Coverage [0.1] & 0.3000 & 0.2000 & 0.4000 & 0.0000 & 0.2000 & 0.1000 & \textbf{0.1000} & 0.1000 \\
    & Coverage [0.5] & 0.3000 & 0.7000 & 0.2000 & 0.6000 & \textbf{0.5000} & 0.4000 & 0.4000 & 0.4000 \\
    & Coverage [0.9] & 0.6000 & 0.7000 & 0.5000 & 0.7000 & 0.8000 & \textbf{0.9000} & 0.8000 & \textbf{0.9000} \\
    & CRPS & 0.4876 & 0.4654 & 0.5123 & 0.4987 & 0.4765 & 0.4321 & \textbf{0.4123} & 0.4234 \\
    & Energy Score & 1.8765 & 1.7898 & 1.9456 & 1.8765 & 1.8321 & 1.5432 & \textbf{1.4987} & 1.5123 \\
    \bottomrule
    \end{tabularx}
    }
    \caption{Comparison of probabilistic forecasting performance across models for 5-day, 7-day, and 10-day horizons on COVID-19 case data, averaged over six countries. Metrics include Quantile Loss at 0.1, 0.5, and 0.9 quantiles, Coverage at corresponding quantiles, Continuous Ranked Probability Score (CRPS), and Energy Score. The proposed STOAT variants (Student-\(t\), Gaussian, Laplace) are compared against baseline models (DeepVAR, DeepAR, DSSM, DeepFactor, MTSNet). Bold values indicate the best performance for each metric and time period.}
    \label{tab:MainResultTransposed}
\end{table*}

\subsubsection{Evaluation Metrics}

% We evaluate STOAT and baseline models using metrics designed to assess probabilistic forecasting performance, focusing on the accuracy of predicted distributions:

We evaluate STOAT and baseline models using metrics designed to assess probabilistic forecasting performance, focusing on the accuracy of predicted distributions. Specifically, we employ the Continuous Ranked Probability Score (CRPS), the Weighted Quantile Loss (WQL), and the Energy Score to quantify how well each model’s predictive distribution aligns with observed outcomes and the details are shown in Appendix~\ref{metrics}.

\subsubsection{Settings}

All experiments are conducted on same GPU to ensure fair comparison. We perform forecasting experiments with prediction horizons of 3, 5, and 7 days, using five different random seeds to ensure robustness. The context window length \(t_0\) and prediction horizon \(m\) are set in a 5:1 ratio (e.g., 15 days for training and 3 days for prediction, 25 days for training and 5 days for prediction, or 35 days for training and 7 days for prediction). Hyperparameters are optimized via a back-test strategy before forecasting. For each prediction task, 100 samples are drawn from the probabilistic decoder to generate forecasts. These settings are consistently applied to STOAT and all baseline models. An ablation study is performed to evaluate the contributions of spatial causal inference (via \(\mathbf{S}\) and \(\mathbf{z}_{it}\)) and causal factors (\(\mathbf{C}\)) to forecasting accuracy.

\subsection{Comparison Results}

\begin{figure}
    \centering
    \includegraphics[width=0.49\textwidth]{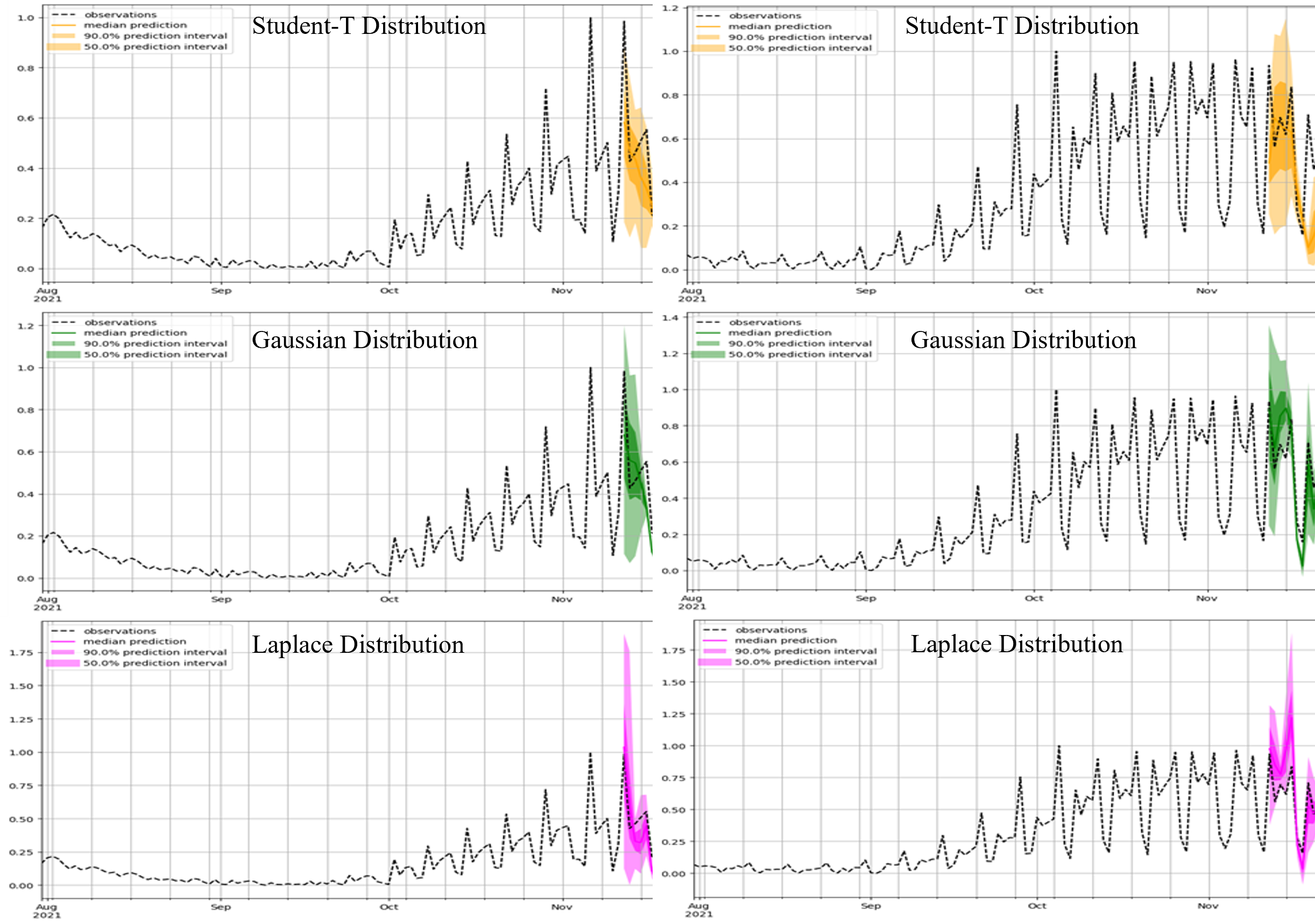}
    \caption{Visualization of 10-day probabilistic forecasts for COVID-19 case data during peak periods in Canada (left) and Italy (right) using the STOAT model. The yellow, green, and red lines represent the predictive distributions based on Student-t, Gaussian, and Laplace distributions, respectively.}
    \vspace{-0.4cm}
    \label{fig:results0}
\end{figure}

This study evaluates probabilistic forecasting performance during the peak phase of the COVID-19 pandemic, focusing on scenarios with significant external causal interventions, using case data from six countries: Canada, France, Italy, UK, US, and Spain. We compare the proposed STOAT model variants (STOAT-StudentT, STOAT-Gaussian, STOAT-Laplace) against baseline models, including DeepVAR, DeepAR, DSSM, DeepFactor, and MTSNet, over 5-day, 7-day, and 10-day horizons, with results averaged across these countries. The experimental outcomes, presented in \tablename{\ref{tab:MainResultTransposed}}, assess performance using Quantile Loss (QL) at 0.1, 0.5, and 0.9 quantiles, Coverage at corresponding quantiles, Continuous Ranked Probability Score (CRPS), and Energy Score. These metrics evaluate distinct aspects of probabilistic forecasts: QL measures accuracy at specific quantiles (lower is better), Coverage assesses calibration by the proportion of true observations within prediction intervals (ideal when matching nominal levels, e.g., 0.9 for 90\% intervals), CRPS quantifies overall distributional accuracy (lower is better), and Energy Score evaluates multivariate forecast quality (lower is better). The results demonstrate STOAT’s superior performance during peak pandemic periods with strong causal interventions, driven by its spatial causal inference and causal factor integration, though its advantages are less pronounced in scenarios with weak causal signals.

The STOAT variants consistently outperform baselines in key metrics, particularly during peak periods. STOAT-Gaussian achieves the lowest CRPS (0.2745, 0.3056, 0.4123 for 5, 7, and 10 days, respectively) and Energy Score (1.0789, 1.1876, 1.4987), indicating superior probabilistic accuracy and multivariate forecast quality. It also excels in QL [0.1] (0.1421, 0.1789, 0.2456), reflecting precise predictions in the lower tail. STOAT-Laplace leads in QL [0.5] and [0.9] for the 10-day horizon (0.6234 and 0.3212), leveraging its robustness to outliers, while STOAT-StudentT achieves the best QL [0.9] for 5 days (0.2356) and matches the optimal Coverage [0.9] (0.9000) across multiple horizons, benefiting from its heavy-tailed distribution to capture extreme case surges. For Coverage, STOAT variants are well-calibrated, achieving ideal values (e.g., 0.1000 for Coverage [0.1], 0.9000 for Coverage [0.9]), unlike DSSM, which shows under-coverage (e.g., 0.6000 for Coverage [0.9] at 5  days), likely due to its Gaussian assumptions failing to model extreme variations driven by interventions like lockdowns.

However, STOAT’s performance is not uniformly superior. DeepAR achieves the lowest QL [0.5] for 5 days (0.3421), and MTSNet matches the optimal Coverage [0.5] (0.5000) for 5 and 7 days, indicating competitive performance in central quantiles. These results highlight the sensitivity of probabilistic forecasting to distributional assumptions and data characteristics across the six countries during peak phases. STOAT’s causal modules (\(\mathbf{Z}_{it}\), \(\mathbf{C}\)) enhance performance by capturing covariates such as mobility restrictions and public health interventions, which are critical during peak periods in countries like the US and Italy. In contrast, baselines like DSSM and DeepFactor, with simpler RNN or state-space architectures, struggle with spatial dependencies, leading to higher QL (e.g., 0.2015 and 0.1976 for QL [0.1] at 5 days) and CRPS (0.3512 and 0.3176). In scenarios with weak causal signals, such as non-peak periods in stable regions like Canada, STOAT variants perform comparably to DeepAR and MTSNet, suggesting that the causal modules’ benefits are context-dependent.

The flexibility of STOAT’s architecture, incorporating Student’s-\(t\), Gaussian, and Laplace distributions, allows it to adapt to heterogeneous time-series patterns across the six countries. For instance, STOAT-Gaussian’s low CRPS reflects its balance of calibration and sharpness, while STOAT-Laplace’s QL performance at higher quantiles suits extreme values. \figurename{\ref{fig:results0}} illustrates STOAT-Laplace’s 10-day forecast for Canada and Italy, showing median predictions and 50\% and 90\% confidence intervals, effectively capturing peak case trends influenced by interventions. However, challenges remain in modeling small, highly variable datasets with cross-country inconsistencies, where no single distribution consistently outperforms across all metrics and horizons. Overall, STOAT’s causal integration enhances its robustness for peak pandemic forecasting under significant external interventions, positioning it as a competitive approach in probabilistic time-series modeling.
\vspace{-0.2cm}

\subsection{Ablation Study}
To evaluate the contributions of spatial causal inference and causal factor integration within STOAT, we perform ablation experiments on COVID-19 case data, averaged over peak periods in six countries. Across three probabilistic output distributions (Laplace, Gaussian, Student-$t$), we compare the full model against ablated variants that remove either the spatial relation matrix or the set of causal covariates.

As shown in Table~\ref{tab:stoat_ablation_full}, the complete STOAT model consistently achieves the best overall performance in terms of CRPS and Energy Score across all forecast horizons. For instance, STOAT-Gaussian achieves a CRPS of 0.2745 and an Energy Score of 1.0789 on the 5-day horizon, outperforming both -NoSpatial and -NoFactors variants. However, we observe that in some cases, ablated models yield lower errors on individual metrics. For example, STOAT-StudentT-NoSpatial attains a lower Quantile Loss [0.1] (0.1512) than the full Student-$t$ variant (0.1543), and STOAT-Gaussian-NoFactors achieves a lower Quantile Loss [0.1] (0.1400) compared to the full Gaussian model (0.1421) on the same horizon. This phenomenon is most pronounced on short-term forecasts, where reduced model complexity or incidental alignment with test-set fluctuations can occasionally result in marginally better pointwise quantile performance.

Nevertheless, as the prediction horizon increases, the benefits of joint spatial and causal modeling become more apparent. For example, on the 10-day forecast, the full STOAT-Gaussian model yields a CRPS of 0.4123, compared to 0.4204 and 0.4229 for the -NoSpatial and -NoFactors ablations, respectively, with a similar trend for the Laplace and Student-$t$ settings. These results highlight that while ablated models may occasionally surpass the full model on isolated quantile losses, the overall probabilistic calibration and sharpness—as captured by proper scoring rules—remain superior for the complete spatial-causal STOAT. This underscores the importance of incorporating both spatial dependencies and causally relevant covariates for robust spatial-temporal probabilistic forecasting.

\begin{table*}
    \centering
    \renewcommand{\arraystretch}{1.0}
    \footnotesize
    \adjustbox{max width=\textwidth}{
    \begin{tabularx}{\textwidth}{
        l|l|
        >{\centering\arraybackslash}X
        >{\centering\arraybackslash}X
        >{\centering\arraybackslash}X|
        >{\centering\arraybackslash}X
        >{\centering\arraybackslash}X
        >{\centering\arraybackslash}X|
        >{\centering\arraybackslash}X
        >{\centering\arraybackslash}X
        >{\centering\arraybackslash}X
    }
    \toprule
    Time Period & Metrics 
    & STOAT-StudentT & STOAT-StudentT-NoSpatial & STOAT-StudentT-NoFactors 
    & STOAT-Gaussian & STOAT-Gaussian-NoSpatial & STOAT-Gaussian-NoFactors 
    & STOAT-Laplace & STOAT-Laplace-NoSpatial & STOAT-Laplace-NoFactors \\
    \midrule
    \multirow{8}{*}{\textbf{5 Days}} 
    & Quantile Loss [0.1] 
        & 0.1543 & \textbf{0.1512} & 0.1585 & 0.1421 & 0.1439 & \textbf{0.1400} & \textbf{0.1478} & 0.1490 & 0.1485 \\
    & Quantile Loss [0.5] 
        & 0.3476 & 0.3487 & \textbf{0.3325} & 0.3378 & \textbf{0.3310} & 0.3383 & \textbf{0.3312} & 0.3325 & 0.3334 \\
    & Quantile Loss [0.9] 
        & \textbf{0.2356} & 0.2370 & 0.2374 & \textbf{0.2412} & 0.2428 & 0.2430 & 0.2389 & \textbf{0.2340} & 0.2416 \\
    & Coverage [0.1] 
        & 0.1000 & 0.1000 & 0.1000 & 0.1000 & 0.1000 & 0.1000 & 0.1000 & 0.1000 & 0.1000 \\
    & Coverage [0.5] 
        & 0.4000 & 0.4000 & 0.4000 & 0.5000 & 0.5000 & 0.5000 & 0.5000 & 0.5000 & 0.5000 \\
    & Coverage [0.9] 
        & 0.9000 & 0.9000 & 0.8000 & 0.8000 & 0.8000 & 0.7000 & 0.9100 & 0.9000 & 0.8000 \\
    & CRPS 
        & \textbf{0.2897} & 0.2920 & 0.2918 & 0.2745 & 0.2770 & \textbf{0.2730} & \textbf{0.2798} & 0.2807 & 0.2812 \\
    & Energy Score 
        & 1.1012 & 1.1035 & \textbf{1.0970} & \textbf{1.0789} & 1.0801 & 1.0812 & 1.0876 & \textbf{1.0871} & 1.0878 \\
    \midrule
    \multirow{8}{*}{\textbf{7 Days}} 
    & Quantile Loss [0.1] 
        & 0.1898 & 0.1932 & \textbf{0.1880} & \textbf{0.1789} & 0.1812 & 0.1831 & 0.1845 & \textbf{0.1825} & 0.1886 \\
    & Quantile Loss [0.5] 
        & 0.4723 & \textbf{0.4690} & 0.4758 & \textbf{0.4612} & 0.4628 & 0.4609 & \textbf{0.4534} & 0.4557 & 0.4565 \\
    & Quantile Loss [0.9] 
        & \textbf{0.2612} & 0.2630 & 0.2638 & \textbf{0.2589} & 0.2599 & 0.2602 & 0.2567 & \textbf{0.2563} & 0.2585 \\
    & Coverage [0.1] 
        & 0.1000 & 0.1000 & 0.1000 & 0.1000 & 0.1000 & 0.1200 & 0.1000 & 0.1000 & 0.1000 \\
    & Coverage [0.5] 
        & 0.4000 & 0.4000 & 0.4000 & 0.5000 & 0.5000 & 0.4000 & 0.5000 & 0.4000 & 0.4000 \\
    & Coverage [0.9] 
        & 0.9000 & 0.8000 & 0.8000 & 0.8000 & 0.8000 & 0.7000 & 0.9000 & 0.9100 & 0.8000 \\
    & CRPS 
        & \textbf{0.3176} & 0.3210 & 0.3225 & \textbf{0.3056} & 0.3079 & 0.3087 & \textbf{0.3123} & 0.3139 & 0.3146 \\
    & Energy Score 
        & 1.2234 & 1.2298 & \textbf{1.2112} & \textbf{1.1876} & 1.1922 & 1.1935 & 1.1987 & \textbf{1.1962} & 1.2029 \\
    \midrule
    \multirow{8}{*}{\textbf{10 Days}} 
    & Quantile Loss [0.1] 
        & \textbf{0.2567} & 0.2685 & 0.2708 & \textbf{0.2456} & 0.2517 & 0.2532 & 0.2512 & 0.2605 & \textbf{0.2450} \\
    & Quantile Loss [0.5] 
        & 0.6654 & 0.6752 & \textbf{0.6385} & \textbf{0.6456} & 0.6564 & 0.6589 & \textbf{0.6234} & 0.6388 & 0.6410 \\
    & Quantile Loss [0.9] 
        & 0.3321 & \textbf{0.3290} & 0.3430 & \textbf{0.3289} & 0.3348 & 0.3364 & \textbf{0.3212} & 0.3281 & 0.3297 \\
    & Coverage [0.1] 
        & 0.1000 & 0.1000 & 0.1000 & 0.1000 & 0.1000 & 0.1000 & 0.1000 & 0.1000 & 0.1000 \\
    & Coverage [0.5] 
        & 0.4000 & 0.4100 & 0.3000 & 0.4000 & 0.4000 & 0.3000 & 0.4000 & 0.3000 & 0.3000 \\
    & Coverage [0.9] 
        & 0.9000 & 0.7000 & 0.7000 & 0.8000 & 0.7000 & 0.7000 & 0.9000 & 0.8000 & 0.8000 \\
    & CRPS 
        & \textbf{0.4321} & 0.4419 & 0.4452 & \textbf{0.4123} & 0.4204 & 0.4229 & \textbf{0.4234} & 0.4341 & 0.4365 \\
    & Energy Score 
        & \textbf{1.5432} & 1.5603 & 1.5632 & \textbf{1.4987} & 1.5199 & 1.5225 & \textbf{1.5123} & 1.5301 & 1.5336 \\
    \bottomrule
    \end{tabularx}
    }
    \caption{Ablation study of STOAT model variants: performance comparison of STOAT-StudentT, STOAT-Gaussian, and STOAT-Laplace distributions, including -NoSpatial and -NoFactors ablations, across 5-day, 7-day, and 10-day horizons on COVID-19 case data. Some individual ablation branches occasionally outperform the full model on specific metrics, yet the full models consistently demonstrate optimal overall performance (CRPS, Energy Score). Metrics in bold indicate the best performance among the STOAT variants.}
    \label{tab:stoat_ablation_full}
\end{table*}

\begin{figure}[!ht]
\centering
\includegraphics[width=0.45\textwidth]{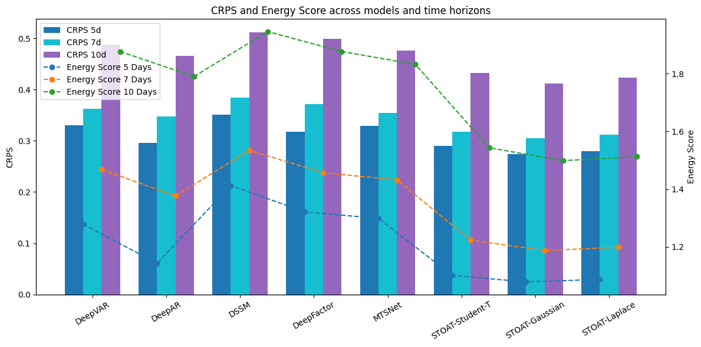}
\vspace{-0.5cm}
\caption{Comparison of probabilistic forecasting models across three forecast horizons (5-day, 7-day, and 10-day) on COVID-19 case data averaged over six countries. The performance is evaluated using the main metrics including CRPS and Energy Score. STOAT variants (Student’s-\(t\), Gaussian, and Laplace) are compared against baseline models (DeepVAR, DeepAR, DSSM, DeepFactor, MTSNet).}
\label{fig:energy_score22}
\end{figure}

\begin{figure}
\centering
\includegraphics[width=0.41\textwidth]{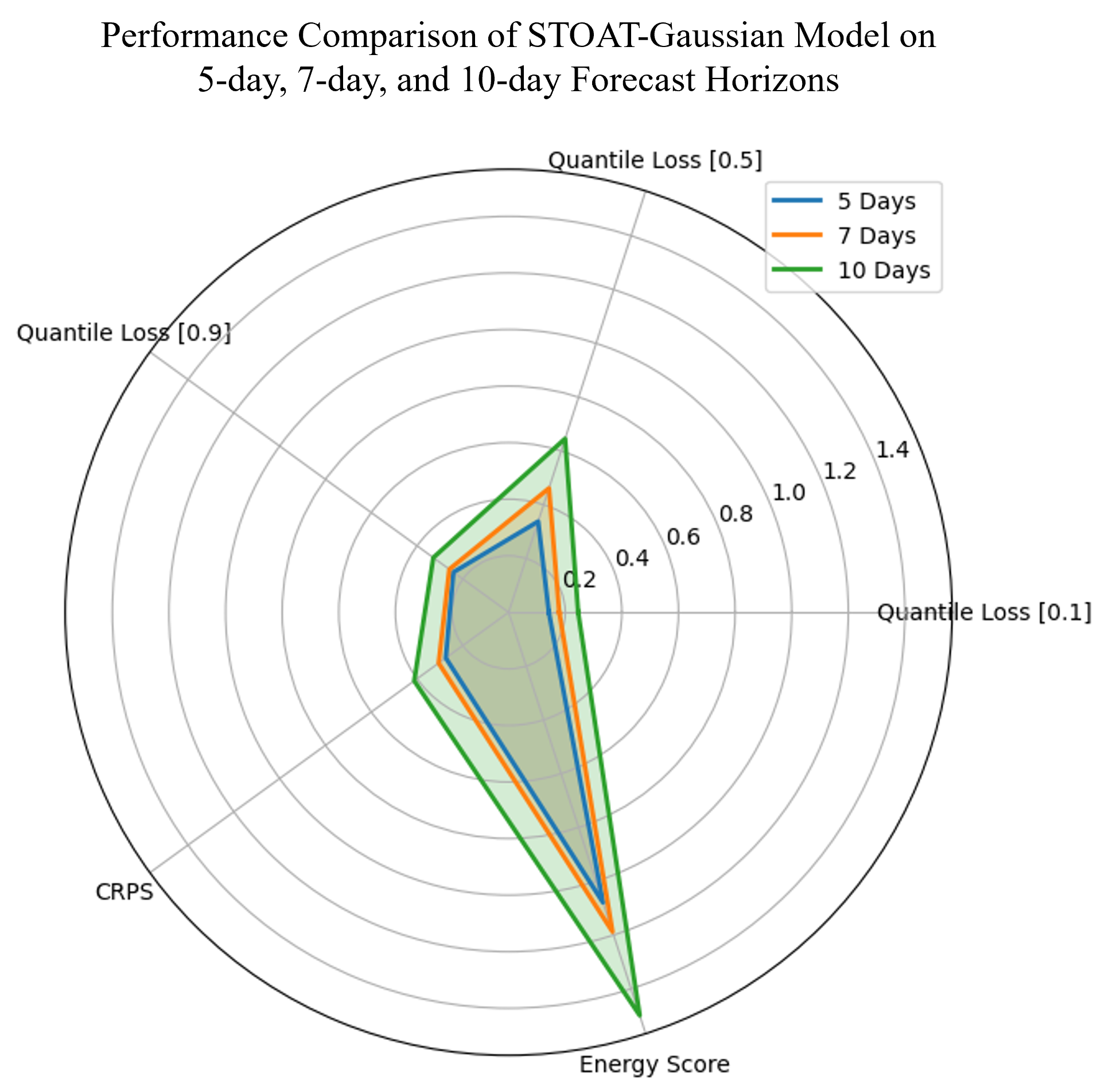}
\caption{
Performance Comparison of STOAT-Gaussian Model on 5-day, 7-day, and 10-day Forecast Horizons.The radar plot demonstrates that the STOAT-Gaussian model performs better on short-term forecasts, with lower Quantile Loss, CRPS, and Energy Score for the 5-day horizon compared to longer horizons, reflecting the expected increase in forecasting uncertainty over time.}
\label{fig:energy_score11}
\end{figure}

\begin{figure}
\centering
\includegraphics[width=0.45\textwidth]{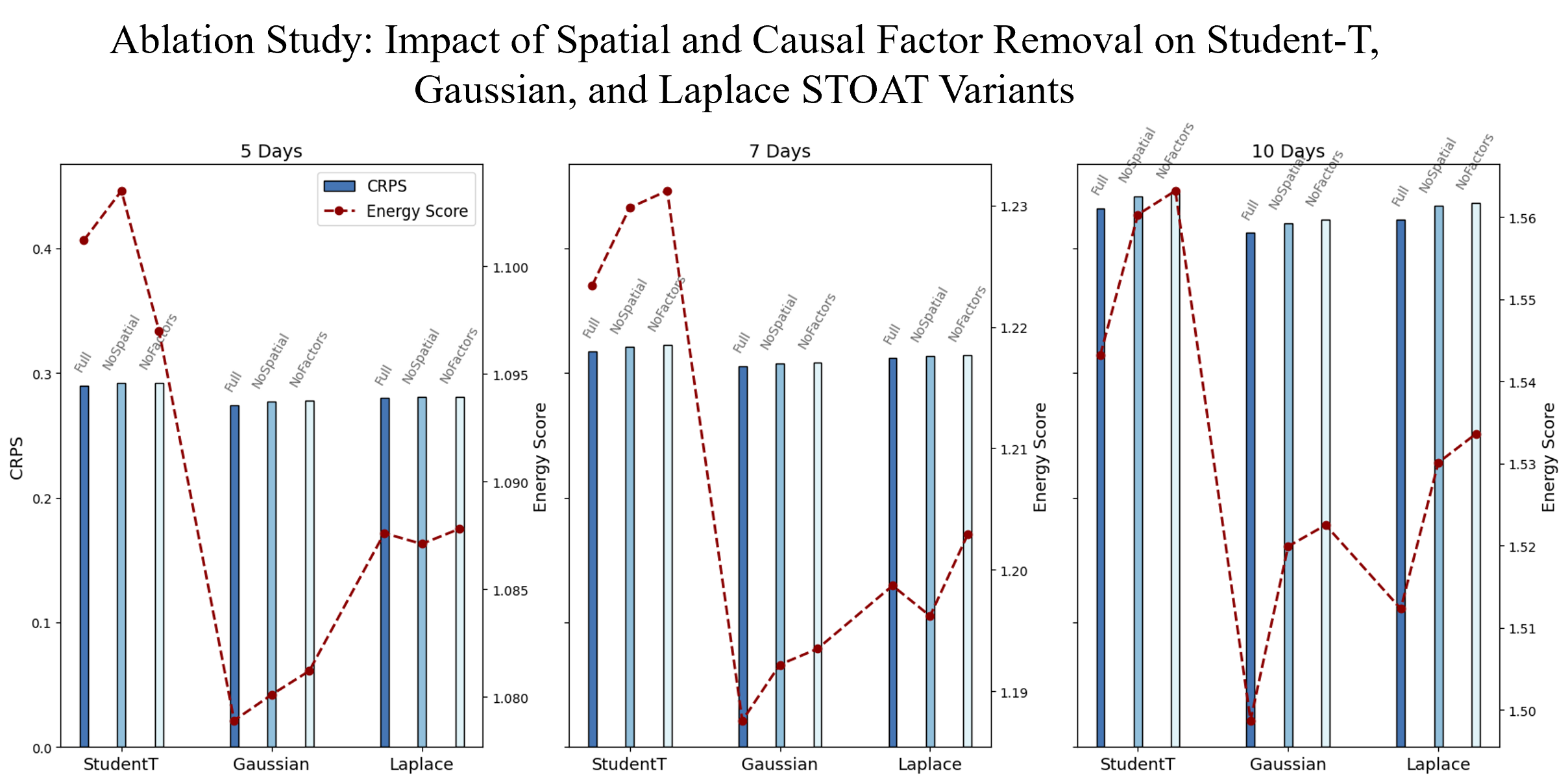}
\caption{
Absolute performance of the STOAT model and its ablation variants across three distribution choices (Student-T, Gaussian, Laplace) and three forecast horizons (5, 7, 10 days). For each distribution, the full model (“Full”) is compared with its “NoSpatial” and “NoFactors” ablation variants, reporting both CRPS (bars, left axis) and Energy Score (lines, right axis). The results show that removing either spatial causal adjustment or causal factor integration generally leads to degraded predictive performance, confirming the contribution of both components to overall model accuracy.}
\vspace{-0.4cm}
\label{fig:Ablation}
\end{figure}

Figure \ref{fig:Ablation} illustrates the Energy Scores for STOAT, STOAT-NoSpatial, STOAT-NoFactors across 3, 5, and 7-day predictions. STOAT achieves the lowest Energy Scores (e.g., 0.294 for 3-day, 0.321 for 7-day), outperforming both its variants and baselines. All models show better performance on shorter horizons, reflecting the challenges of long-term forecasting for volatile STC-TS data. Notably, STOAT-NoSpatial outperforms STOAT-NoFactors by 2.5\% in CRPS for 7-day predictions, suggesting that spatial causal inference (via \(\mathbf{S}\) ) contributes more to long-term accuracy than causal factors alone. These findings highlight the synergistic effect of STOAT’s modules in robust probabilistic forecasting.

\subsection{Conclusion}

Spatial-temporal causal time series (STC-TS) are prevalent in domains like epidemiology, where region-specific observations are driven by causally relevant covariates and interconnected through spatial dependencies. Existing deep probabilistic methods often treat covariates as exogenous inputs, neglecting spatial causal relationships and producing miscalibrated forecasts. We propose STOAT, a novel framework that addresses these limitations through three key contributions: (1) spatial causal adjustment via extended Difference-in-Differences with learnable spatial relation matrices, (2) causal representation learning that constructs causally adjusted latent representations reflecting spatial spillovers and intervention effects, and (3) causally informed probabilistic forecasting with flexible output distributions conditioned on causally disentangled dynamics. Experiments on COVID-19 data from six countries demonstrate STOAT's superior performance, with STOAT-Gaussian achieving the lowest CRPS and Energy Score across forecast horizons, outperforming state-of-the-art deep probabilistic forecasting baselines including DeepAR, DSSM, DeepFactor, MTSNet, and DeepVAR. Ablation studies confirm the synergistic effects of spatial causal inference and causal factors. However, STOAT assumes static spatial correlations and shows diminished benefits in scenarios with weak causal signals. Future work will focus on dynamic spatial modeling, mixture models for heterogeneous dependencies, and real-time causal intervention integration for broader STC-TS applications.

%\section{Appendices}

%%``\verb|\end{document}|'' command at the conclusion of your source document.

%% The acknowledgments section is defined using the "acks" environment
%% (and NOT an unnumbered section). This ensures the proper
%% identification of the section in the article metadata, and the
%% consistent spelling of the heading.
%%
%% The next two lines define the bibliography style to be used, and
%% the bibliography file.
\newpage
\bibliographystyle{ACM-Reference-Format}
\bibliography{sample-base}

@String{Computing = "Computing" }

@String{Springer = "Springer-Verlag" }

@article{cao2022ai,
  title={AI in combating the COVID-19 pandemic},
  author={Cao, Longbing},
  journal={IEEE Intelligent Systems},
  volume={37},
  number={2},
  pages={3--13},
  year={2022},
  publisher={IEEE}
}

@article{cao2021covid,
	author = {Cao, Longbing and Liu, Qing},
	title = {COVID-19 Modeling: A Review},
	year = {2022},
    pages = {1-105},
	doi = {10.1101/2022.08.22.22279022},
	URL = {https://www.medrxiv.org/content/early/2022/09/26/2022.08.22.22279022},
	journal = {medRxiv}
}

@article{mastakouri2020causal,
  title={Causal analysis of Covid-19 spread in Germany},
  author={Mastakouri, Atalanti and Sch{\"o}lkopf, Bernhard},
  journal={Advances in Neural Information Processing Systems},
  volume={33},
  pages={3153--3163},
  year={2020}
}

@article{nolan2020exploring,
  title={Exploring the impact of COVID-19 lockdown on social roles and emotions while working from home},
  author={Nolan, Sam and Rumi, Shakila Khan and Anderson, Christoph and David, Klaus and Salim, Flora D},
  journal={arXiv preprint arXiv:2007.12353},
  year={2020}
}

@article{wang2024learning,
  title={Learning spatio-temporal dynamics on mobility networks for adaptation to open-world events},
  author={Wang, Zhaonan and Jiang, Renhe and Xue, Hao and Salim, Flora D and Song, Xuan and Shibasaki, Ryosuke and Hu, Wei and Wang, Shaowen},
  journal={Artificial Intelligence},
  volume={335},
  pages={104120},
  year={2024},
  publisher={Elsevier}
}

@inproceedings{salim2021learning,
  title={Learning Spatio-Temporal Behavioural Representations for Urban Activity Forecasting},
  author={Salim, Flora D.},
  booktitle={Companion Proceedings of The Web Conference 2021},
  publisher={Association for Computing Machinery},
  address={New York, NY, USA},
  pages={347--348},
  year={2021},
  doi={10.1145/3442442.3451892},
  url={https://dl.acm.org/doi/abs/10.1145/3442442.3451892}
}

@article{du2025blue,
  title={BLUE: Bi-layer Heterogeneous Graph Fusion Network for Avian Influenza Forecasting},
  author={Du, Jing and Stone, Haley and Yang, Yang and Desai, Ashna and Xue, Hao and Z{\"u}fle, Andreas and MacIntyre, Chandini Raina and Salim, Flora D},
  journal={arXiv preprint arXiv:2505.22692},
  year={2025}
}

@article{salinas2020deepar,
  title={DeepAR: Probabilistic forecasting with autoregressive recurrent networks},
  author={Salinas, David and Flunkert, Valentin and Gasthaus, Jan and Januschowski, Tim},
  journal={International journal of forecasting},
  volume={36},
  number={3},
  pages={1181--1191},
  year={2020},
  publisher={Elsevier}
}

@article{rangapuram2018deep,
  title={Deep state space models for time series forecasting},
  author={Rangapuram, Syama Sundar and Seeger, Matthias W and Gasthaus, Jan and Stella, Lorenzo and Wang, Yuyang and Januschowski, Tim},
  journal={Advances in neural information processing systems},
  volume={31},
  year={2018}
}

@article{salinas2019high,
  title={High-dimensional multivariate forecasting with low-rank gaussian copula processes},
  author={Salinas, David and Bohlke-Schneider, Michael and Callot, Laurent and Medico, Roberto and Gasthaus, Jan},
  journal={Advances in neural information processing systems},
  volume={32},
  year={2019}
}

@article{metcalfe2019pay,
  title={Pay for performance and hip fracture outcomes an interrupted time series and difference-in-differences analysis in England and Scotland},
  author={Metcalfe, David and Zogg, Cheryl K and Judge, Andrew and Perry, Daniel C and Gabbe, B and Willett, K and Costa, ML},
  journal={The bone \& joint journal},
  volume={101},
  number={8},
  pages={1015--1023},
  year={2019},
  publisher={Bone \& Joint}
}

@article{rothbard2024tutorial,
  title={A tutorial on applying the difference-in-differences method to health data},
  author={Rothbard, Sarah and Etheridge, James C and Murray, Eleanor J},
  journal={Current Epidemiology Reports},
  volume={11},
  number={2},
  pages={85--95},
  year={2024},
  publisher={Springer}
}

@article{ryan2019now,
  title={Now trending: Coping with non-parallel trends in difference-in-differences analysis},
  author={Ryan, Andrew M and Kontopantelis, Evangelos and Linden, Ariel and Burgess Jr, James F},
  journal={Statistical methods in medical research},
  volume={28},
  number={12},
  pages={3697--3711},
  year={2019},
  publisher={SAGE Publications Sage UK: London, England}
}

@article{naumann2020covid,
  title={COVID-19 policies in Germany and their social, political, and psychological consequences},
  author={Naumann, Elias and M{\"o}hring, Katja and Reifenscheid, Maximiliane and Wenz, Alexander and Rettig, Tobias and Lehrer, Roni and Krieger, Ulrich and Juhl, Sebastian and Friedel, Sabine and Fikel, Marina and others},
  journal={European Policy Analysis},
  volume={6},
  number={2},
  pages={191--202},
  year={2020},
  publisher={Wiley Online Library}
}

@article{cooper2020sir,
  title={A SIR model assumption for the spread of COVID-19 in different communities},
  author={Cooper, Ian and Mondal, Argha and Antonopoulos, Chris G},
  journal={Chaos, Solitons \& Fractals},
  volume={139},
  pages={110057},
  year={2020},
  publisher={Elsevier}
}

@article{tchetgen2024universal,
  title={Universal difference-in-differences for causal inference in epidemiology},
  author={Tchetgen, Eric J Tchetgen and Park, Chan and Richardson, David B},
  journal={Epidemiology},
  volume={35},
  number={1},
  pages={16--22},
  year={2024},
  publisher={LWW}
}

@article{ohlsson2020applying,
  title={Applying causal inference methods in psychiatric epidemiology: A review},
  author={Ohlsson, Henrik and Kendler, Kenneth S},
  journal={JAMA psychiatry},
  volume={77},
  number={6},
  pages={637--644},
  year={2020},
  publisher={American Medical Association}
}

@article{kane2020propensity,
  title={Propensity score matching: a statistical method},
  author={Kane, Liam T and Fang, Taolin and Galetta, Matthew S and Goyal, Dhruv KC and Nicholson, Kristen J and Kepler, Christopher K and Vaccaro, Alexander R and Schroeder, Gregory D},
  journal={Clinical spine surgery},
  volume={33},
  number={3},
  pages={120--122},
  year={2020},
  publisher={LWW}
}

@article{baiocchi2014instrumental,
  title={Instrumental variable methods for causal inference},
  author={Baiocchi, Michael and Cheng, Jing and Small, Dylan S},
  journal={Statistics in medicine},
  volume={33},
  number={13},
  pages={2297--2340},
  year={2014},
  publisher={Wiley Online Library}
}

@article{xie2023overview,
  title={An overview of deterministic and probabilistic forecasting methods of wind energy},
  author={Xie, Yuying and Li, Chaoshun and Li, Mengying and Liu, Fangjie and Taukenova, Meruyert},
  journal={Iscience},
  volume={26},
  number={1},
  year={2023},
  publisher={Elsevier}
}

@article{gneiting2014probabilistic,
  title={Probabilistic forecasting},
  author={Gneiting, Tilmann and Katzfuss, Matthias},
  journal={Annual Review of Statistics and Its Application},
  volume={1},
  number={1},
  pages={125--151},
  year={2014},
  publisher={Annual Reviews}
}

@article{jordan2019evaluating,
  title={Evaluating probabilistic forecasts with scoringRules},
  author={Jordan, Alexander and Kr{\"u}ger, Fabian and Lerch, Sebastian},
  journal={Journal of Statistical Software},
  volume={90},
  pages={1--37},
  year={2019}
}

@article{gneiting2005calibrated,
  title={Calibrated probabilistic forecasting using ensemble model output statistics and minimum CRPS estimation},
  author={Gneiting, Tilmann and Raftery, Adrian E and Westveld III, Anton H and Goldman, Tom},
  journal={Monthly weather review},
  volume={133},
  number={5},
  pages={1098--1118},
  year={2005}
}

@article{hale2021global,
  title={A global panel database of pandemic policies (Oxford COVID-19 Government Response Tracker)},
  author={Hale, Thomas and Angrist, Noam and Goldszmidt, Rafael and Kira, Beatriz and Petherick, Anna and Phillips, Toby and Webster, Samuel and Cameron-Blake, Emily and Hallas, Laura and Majumdar, Saptarshi and others},
  journal={Nature human behaviour},
  volume={5},
  number={4},
  pages={529--538},
  year={2021},
  publisher={Nature Publishing Group UK London}
}

@inproceedings{wang2019deep,
  title={Deep factors for forecasting},
  author={Wang, Yuyang and Smola, Alex and Maddix, Danielle and Gasthaus, Jan and Foster, Dean and Januschowski, Tim},
  booktitle={International conference on machine learning},
  pages={6607--6617},
  year={2019},
  organization={PMLR},
  publisher={Proceedings of Machine Learning Research (PMLR)},
  address={Long Beach, California, USA}
}

@article{abadie2005semiparametric,
  title={Semiparametric difference-in-differences estimators},
  author={Abadie, Alberto},
  journal={The review of economic studies},
  volume={72},
  number={1},
  pages={1--19},
  year={2005},
  publisher={Wiley-Blackwell}
}

@article{moraffah2021causal,
  title={Causal inference for time series analysis: Problems, methods and evaluation},
  author={Moraffah, Raha and Sheth, Paras and Karami, Mansooreh and Bhattacharya, Anchit and Wang, Qianru and Tahir, Anique and Raglin, Adrienne and Liu, Huan},
  journal={Knowledge and Information Systems},
  volume={63},
  pages={3041--3085},
  year={2021},
  publisher={Springer}
}

@article{athey2017state,
  title={The state of applied econometrics: Causality and policy evaluation},
  author={Athey, Susan and Imbens, Guido W},
  journal={Journal of Economic perspectives},
  volume={31},
  number={2},
  pages={3--32},
  year={2017},
  publisher={American Economic Association 2014 Broadway, Suite 305, Nashville, TN 37203-2418}
}

@article{sun2021estimating,
  title={Estimating dynamic treatment effects in event studies with heterogeneous treatment effects},
  author={Sun, Liyang and Abraham, Sarah},
  journal={Journal of econometrics},
  volume={225},
  number={2},
  pages={175--199},
  year={2021},
  publisher={Elsevier}
}

@article{abadie2010synthetic,
  title={Synthetic control methods for comparative case studies: Estimating the effect of California’s tobacco control program},
  author={Abadie, Alberto and Diamond, Alexis and Hainmueller, Jens},
  journal={Journal of the American statistical Association},
  volume={105},
  number={490},
  pages={493--505},
  year={2010},
  publisher={Taylor \& Francis}
}

@article{zhou2024machine,
  title={Machine learning-based causal inference for evaluating intervention in travel behaviour research: A difference-in-differences framework},
  author={Zhou, Meng and Huang, Sixian and Tu, Wei and Wang, Donggen},
  journal={Travel Behaviour and Society},
  volume={37},
  pages={100852},
  year={2024},
  publisher={Elsevier}
}

@inproceedings{schwab2020learning,
  title={Learning counterfactual representations for estimating individual dose-response curves},
  author={Schwab, Patrick and Linhardt, Lorenz and Bauer, Stefan and Buhmann, Joachim M and Karlen, Walter},
  booktitle={Proceedings of the AAAI Conference on Artificial Intelligence},
  volume={34},
  number={04},
  pages={5612--5619},
  year={2020}
}

@inproceedings{yang2023mtsnet,
  title={MTSNet: Deep probabilistic cross-multivariate time series modeling with external factors for COVID-19},
  author={Yang, Yang and Cao, Longbing},
  booktitle={2023 International Joint Conference on Neural Networks (IJCNN)},
  pages={1--10},
  year={2023},
  organization={IEEE},
  publisher={IEEE}
}

@article{alexandrov2020gluonts,
  title={Gluonts: Probabilistic and neural time series modeling in python},
  author={Alexandrov, Alexander and Benidis, Konstantinos and Bohlke-Schneider, Michael and Flunkert, Valentin and Gasthaus, Jan and Januschowski, Tim and Maddix, Danielle C and Rangapuram, Syama and Salinas, David and Schulz, Jasper and others},
  journal={Journal of Machine Learning Research},
  volume={21},
  number={116},
  pages={1--6},
  year={2020}
}

@article{pathan2020time,
  title={Time series prediction of COVID-19 by mutation rate analysis using recurrent neural network-based LSTM model},
  author={Pathan, Refat Khan and Biswas, Munmun and Khandaker, Mayeen Uddin},
  journal={Chaos, Solitons \& Fractals},
  volume={138},
  pages={110018},
  year={2020},
  publisher={Elsevier}
}

@article{alassafi2022time,
  title={Time series predicting of COVID-19 based on deep learning},
  author={Alassafi, Madini O and Jarrah, Mutasem and Alotaibi, Reem},
  journal={Neurocomputing},
  volume={468},
  pages={335--344},
  year={2022},
  publisher={Elsevier}
}

@article{shahid2020predictions,
  title={Predictions for COVID-19 with deep learning models of LSTM, GRU and Bi-LSTM},
  author={Shahid, Farah and Zameer, Aneela and Muneeb, Muhammad},
  journal={Chaos, Solitons \& Fractals},
  volume={140},
  pages={110212},
  year={2020},
  publisher={Elsevier}
}

@article{he2020seir,
  title={SEIR modeling of the COVID-19 and its dynamics},
  author={He, Shaobo and Peng, Yuexi and Sun, Kehui},
  journal={Nonlinear dynamics},
  volume={101},
  pages={1667--1680},
  year={2020},
  publisher={Springer}
}

@article{carcione2020simulation,
  title={A simulation of a COVID-19 epidemic based on a deterministic SEIR model},
  author={Carcione, Jos{\'e} M and Santos, Juan E and Bagaini, Claudio and Ba, Jing},
  journal={Frontiers in public health},
  volume={8},
  pages={230},
  year={2020},
  publisher={Frontiers Media SA}
}

@article{lopez2021modified,
  title={A modified SEIR model to predict the COVID-19 outbreak in Spain and Italy: simulating control scenarios and multi-scale epidemics},
  author={L{\'o}pez, Leonardo and Rodo, Xavier},
  journal={Results in physics},
  volume={21},
  pages={103746},
  year={2021},
  publisher={Elsevier}
}

@article{runge2023causal,
  title={Causal inference for time series},
  author={Runge, Jakob and Gerhardus, Andreas and Varando, Gherardo and Eyring, Veronika and Camps-Valls, Gustau},
  journal={Nature Reviews Earth \& Environment},
  volume={4},
  number={7},
  pages={487--505},
  year={2023},
  publisher={Nature Publishing Group UK London}
}

@article{gros2021new,
  title={A new COVID policy stringency index for Europe},
  author={Gros, Daniel and Ounnas, Alexandre and Yeung, Timothy Yu-Cheong},
  journal={Covid Economics},
  volume={115},
  year={2021}
}

@article{alimohamadi2020estimate,
  title={Estimate of the basic reproduction number for COVID-19: a systematic review and meta-analysis},
  author={Alimohamadi, Yousef and Taghdir, Maryam and Sepandi, Mojtaba},
  journal={Journal of Preventive Medicine and Public Health},
  volume={53},
  number={3},
  pages={151},
  year={2020}
}

\newpage
\appendix
\section{Appendix}

\subsection{Algorithm}
\label{Algorithm}

\begin{algorithm}
\caption{STOAT Framework: Spatial-Temporal Causal Inference and Probabilistic Forecasting}
\KwIn{%
    Target time series \(\mathbf{y} \in \mathbb{R}^{N \times T}\),
    Covariates \(\mathbf{c} \in \mathbb{R}^{N \times T \times D}\),
    Intervention indicator \(\mathbf{T} \in \{0,1\}^{N \times T}\),
    Post-intervention indicator \(\text{Post}_t \in \{0,1\}^T\),
    Spatial coordinates \(\{(\text{lat}_i, \text{lon}_i)\}_{i=1}^N\),
    Lookback window size \(L\)
}
\KwOut{%
    Probabilistic forecast parameters \(\boldsymbol{\Theta}\),
    Causal effect \(\delta\),
    Spatial effect \(\rho\),
    Covariate effects \(\boldsymbol{\gamma}\)
}
\textbf{Step 1: Constructing spatial relation matrix} \\
\For{$i \gets 1$ \KwTo $N$}{
    \For{$j \gets 1$ \KwTo $N$}{
        \If{$i \neq j$}{
            \(d_{i,j} \gets \texttt{GeodesicDistance}(\text{lat}_i, \text{lon}_i, \text{lat}_j, \text{lon}_j)\) \\
            \(S_{i,j} \gets \frac{1}{d_{i,j}^\alpha}\) 
        }
        \Else{
            \(S_{i,j} \gets 0\) 
        }
    }
    Normalize: \(S_{i\cdot} \gets \frac{S_{i\cdot}}{\sum_{j} S_{i,j}}\) 
}
\textbf{Step 2: Computing spatial lag} \\
\For{$i \gets 1$ \KwTo $N$}{
    \For{$t \gets 1$ \KwTo $T$}{
        \(\text{Lag}_{i,t} \gets \sum_{j=1}^N S_{i,j} y_{j,t}\)
    }
}
\textbf{Step 3: Estimating causal and spatial effects} \\
Fit spatial DiD model (Eq. (\ref{eq:spatial_did})): \\
Estimate \(\hat{\rho}\), \(\hat{\delta}\), \(\boldsymbol{\gamma}\) using instrumental variables and OLS \tcp*{Obtain causal effect \(\hat{\delta}\), spatial effect \(\hat{\rho}\), covariate effects \(\boldsymbol{\gamma}\)} 
\textbf{Step 4: Causal adjustment} \\
\For{$i \gets 1$ \KwTo $N$}{
    \For{$t \gets 1$ \KwTo $T$}{
        \(\tilde{y}_{i,t} \gets y_{i,t} - \hat{\delta} \cdot (T_i \cdot \text{Post}_t)\)
        \(\mathbf{z}_{i,t} \gets \tilde{y}_{i,t} + \hat{\rho} \sum_{j=1}^{N} S_{i,j} \tilde{y}_{j,t}\) \tcp*{Causally adjusted representation (Eq. (\ref{eq:spatial_adjusted_input}))}
    }
}
\textbf{Step 5: Preparing inputs for probabilistic forecasting} \\
\For{$i \gets 1$ \KwTo $N$}{
    \For{$t \gets L$ \KwTo $T-1$}{
        Prepare inputs \(\mathbf{z}_{i,t-L:t}\), \(\mathbf{y}_{i,t-L:t}\) \tcp*{Lookback window \(L\) for causally adjusted and target series}
        Set target \(y_{i,t+1} \gets \tilde{y}_{i,t+1}\) \tcp*{Target for next time step}
    }
}
\textbf{Step 6: Training deep probabilistic model} \\
\textbf{Step 7: Generating probabilistic forecasts} \\
\KwRet{predicted distribution P(.)}
\end{algorithm}

\subsection{Metrics}
\label{metrics}
\begin{itemize}
    \item \textit{Continuous Ranked Probability Score (CRPS) \cite{gneiting2005calibrated}}: Quantifies the discrepancy between predicted and empirical cumulative distribution functions:
    \begin{equation}
        \operatorname{CRPS}(P, x_{i,t}) = \int_{-\infty}^{\infty} \left( P(y) - \mathbf{1}\{y \geq x_{i,t}\} \right)^2 \mathrm{d}y,
        \label{eq:crps}
    \end{equation}
    where \(P\) is the predicted cumulative distribution, and \(\mathbf{1}\{y \geq x_{i,t}\}\) is the indicator function for observed values \(x_{i,t}\). CRPS evaluates the overall quality of probabilistic forecasts by measuring how well the predicted distribution matches the observed values. \textbf{Lower CRPS values indicate better performance}, with perfect forecasts achieving CRPS = 0.
    \item \textit{Weighted Quantile Loss (WQL)}: Measures accuracy at specific quantiles of the predicted distribution:
    \begin{equation}
        \begin{aligned}
            \text{WQL}[\tau] &= 2 \frac{\sum_{i,t} Q_{i,t}^{(\tau)}}{\sum_{i,t} |x_{i,t}|}, \\
            Q_{i,t}^{(\tau)} &= \begin{cases} 
                (1-\tau) |q_{i,t}^{(\tau)} - x_{i,t}| & \text{if } q_{i,t}^{(\tau)} > x_{i,t}, \\
                \tau |q_{i,t}^{(\tau)} - x_{i,t}| & \text{otherwise},
            \end{cases}
        \end{aligned}
        \label{eq:wql}
    \end{equation}
    where \(\tau \in \{0.1, 0.5, 0.9\}\) represents quantiles, and \(q_{i,t}^{(\tau)}\) is the predicted value at quantile \(\tau\). WQL assesses the precision of quantile predictions, with asymmetric penalties that reflect the importance of over- and under-estimation. \textbf{Lower WQL values indicate better quantile forecast accuracy}.
    \item \textit{Coverage}: Measures the proportion of observations that fall within predicted confidence intervals:
    \begin{equation}
        \text{Coverage}[\alpha] = \frac{1}{N} \sum_{i,t} \mathbf{1}\left\{q_{i,t}^{(\alpha/2)} \leq x_{i,t} \leq q_{i,t}^{(1-\alpha/2)}\right\},
        \label{eq:coverage}
    \end{equation}
    where \(N\) is the total number of observations, \(\alpha\) is the significance level (e.g., \(\alpha = 0.1\) for 90\% intervals), and \(q_{i,t}^{(\cdot)}\) are the predicted quantiles. \textbf{The best performance occurs when Coverage matches the nominal level} (i.e., Coverage[0.1] \(\approx 0.9\) and Coverage[0.5] \(\approx 0.5\)), indicating well-calibrated prediction intervals.
    \item \textit{Energy Score} \cite{ jordan2019evaluating}: Evaluates multivariate probabilistic forecast accuracy, sensitive to inter-variable correlations:
    \begin{equation}
        \begin{aligned}
            \operatorname{Energy}(\{\tilde{x}_t, x_t\}) &= \mathbb{E}_{\tilde{x}_t} \|\tilde{x}_t - x_t\|_{Fro}^\beta \\
            &\quad - \frac{1}{2} \mathbb{E}_{\tilde{x}_t, \tilde{x}_t'} \|\tilde{x}_t - \tilde{x}_t'\|_{Fro}^\beta,
        \end{aligned}
        \label{eq:energy}
    \end{equation}
    where \(\tilde{x}_t\) and \(\tilde{x}_t'\) are independent samples from the predicted distribution, \(x_t\) is the observed value, \(\beta = 1\), and \(\|\cdot\|_{Fro}\) denotes the Frobenius norm. Energy Score captures both the accuracy of multivariate forecasts and the quality of predicted dependence structures between variables. \textbf{Lower Energy Score values indicate better multivariate forecast performance}, with the metric being particularly valuable for high-dimensional time series.
\end{itemize}

\end{document}